\documentclass{article}

    \PassOptionsToPackage{numbers, compress}{natbib}

\usepackage[main, final]{neurips_2026}

\usepackage[utf8]{inputenc} 
\usepackage[T1]{fontenc}    
\usepackage{hyperref}       
\usepackage{url}            
\usepackage{booktabs}       
\usepackage{amsfonts}       
\usepackage{nicefrac}       
\usepackage{microtype}      
\usepackage{xcolor}         
\usepackage{placeins}       
\usepackage{graphicx}       
\usepackage{subcaption}     
\usepackage{amsmath}
\usepackage{amssymb}
\usepackage{algorithm}
\usepackage[noend]{algpseudocode}
\usepackage{mathrsfs}
\usepackage{dsfont}
\usepackage{array}
\usepackage{bbm}
\usepackage[mathscr]{eucal}
\usepackage{psfrag}

\usepackage{wasysym}
\usepackage[dvipsnames, table]{xcolor}  
\usepackage{caption}                     
\usepackage{subcaption}
\usepackage{amsthm}
\usepackage{todonotes}
\usepackage{tabulary}
\usepackage{outlines}
\usepackage{thmtools, thm-restate}

\newtheorem{proposition} {Proposition}
\newtheorem{lemma} {Lemma}

\newtheorem{remark} {Remark}

\definecolor{ian_highlight}{RGB}{100, 2, 2}

\errorcontextlines\maxdimen

\makeatletter
    \newcommand*{\algrule}[1][\algorithmicindent]{\makebox[#1][l]{\hspace*{.5em}\thealgruleextra\vrule height \thealgruleheight depth \thealgruledepth}}%
\newcommand*{\thealgruleextra}{}
\newcommand*{\thealgruleheight}{.75\baselineskip}
\newcommand*{\thealgruledepth}{.25\baselineskip}

\newcount\ALG@printindent@tempcnta
\def\ALG@printindent{%
    \ifnum \theALG@nested>0
        \ifx\ALG@text\ALG@x@notext
        \else
            \unskip
            \addvspace{-1pt}
            \ALG@printindent@tempcnta=1
            \loop
                \algrule[\csname ALG@ind@\the\ALG@printindent@tempcnta\endcsname]%
                \advance \ALG@printindent@tempcnta 1
            \ifnum \ALG@printindent@tempcnta<\numexpr\theALG@nested+1\relax
            \repeat
        \fi
    \fi
    }%
\usepackage{etoolbox}
\makeatother

\newbox\statebox
\newcommand{\myState}[1]{%
    \setbox\statebox=\vbox{#1}%
    \edef\thealgruleheight{\dimexpr \the\ht\statebox+1pt\relax}%
    \edef\thealgruledepth{\dimexpr \the\dp\statebox+1pt\relax}%
    \ifdim\thealgruleheight<.75\baselineskip
        \def\thealgruleheight{\dimexpr .75\baselineskip+1pt\relax}%
    \fi
    \ifdim\thealgruledepth<.25\baselineskip
        \def\thealgruledepth{\dimexpr .25\baselineskip+1pt\relax}%
    \fi
    \State #1%
    \def\thealgruleheight{\dimexpr .75\baselineskip+1pt\relax}%
    \def\thealgruledepth{\dimexpr .25\baselineskip+1pt\relax}%
}

\newcommand{\codeurl}{\url{https://github.com/google-deepmind/egg}}

\newif\ifshowacks
\showackstrue  

\title{Delightful Policy Gradient under Distributed Friction}

\author{%
  Ian Osband \\
  Google DeepMind \\
  \texttt{iosband@google.com} \\
}

\begin{document}

\maketitle
\vspace{-1mm}
\begin{abstract}
\vspace{-1mm}
Distributed reinforcement learning trains on data from stale, buggy, or mismatched actors, whose actions often have high surprisal (negative log-probability) under the learner's policy.
Supervised fine-tuning learns stably from such logged data, while policy gradient often does not.
We trace the difference to \emph{negative learning from surprising data}: rare failures can dominate finite-batch updates, while rare successes carry information the current policy would otherwise miss.
The \textit{Delightful Policy Gradient} (DG) separates the two by gating each term with delight, the product of advantage and surprisal, and needs no behavior probabilities.
In a contaminated bandit, DG suppresses the perpendicular noise of rare failures by a factor that vanishes as the learner improves; no gate based only on learner probability can do this without also suppressing rare successes.
On MNIST with stale actors, DG beats importance-weighted policy gradient given the exact behavior probabilities.
On a transformer sequence task, DG often reaches an order of magnitude lower error under staleness, actor--learner probability mismatch, actor bugs, reward corruption, and rare discovery.
With four frictions combined, its sample-efficiency advantage grows with sequence length.
All experiments are controlled diagnostics; we do not evaluate an end-to-end distributed system.
\end{abstract}

\vspace{-1mm}
\section{Introduction}
\label{sec:intro}
\vspace{-1mm}

Large-scale post-training for reasoning models runs policy gradient through distributed stacks~\citep{openai2024o1,guo2025deepseek}.
Rollouts and gradients come from different backends, actor versions, or inference implementations.
Even identical weights can assign different token probabilities across these systems, and small mismatches compound across tokens, silently turning on-policy training into off-policy training~\citep{yao2025offpolicy,he2025nondeterminism}.
Supervised fine-tuning trains on the same kind of logged data from the same distributed actors, yet it is far more stable and needs no behavior probabilities.

SFT only increases the log-probability of observed targets, whereas policy gradient also applies negative updates.
Stale actors, buggy implementations, and mismatched inference stacks all generate actions that surprise the learner's current policy, and many of these actions fail.
We call the resulting failure mode \emph{negative learning from surprising data}: high-surprisal failures can dominate a finite batch through large components perpendicular to the useful direction, even when what they teach is redundant.
Existing fixes reconstruct or stabilize the policy gradient: importance weighting corrects for actor--learner differences when behavior probabilities are known~\citep{precup2001off}, and TRPO and PPO constrain unstable updates~\citep{schulman2015trust,schulman2017proximal}, but neither distinguishes surprising failures from surprising successes.

The Delightful Policy Gradient (DG) separates the two by gating each update with \emph{delight}, the product of advantage and action surprisal, i.e.\ the negative log-probability under the current policy~\citep{osband2026delightfulpolicygradient}.
When a surprising action succeeds, a \emph{breakthrough}, the gate opens and the update is kept; when it fails, a \emph{blunder}, the gate closes and the update is suppressed.
Because surprisal is computed under the learner's current policy, DG requires no behavior probabilities and no knowledge of the friction source.

The update rule and the rare-success versus rare-failure intuition come from \citet{osband2026delightfulpolicygradient}.
This paper asks whether those gains survive contaminated data, and finds that DG's advantage grows with the level of contamination.
On MNIST with stale actors, DG outperforms importance weighting with exact behavior probabilities, so the reconstructed on-policy gradient is not the best finite-batch update (Section~\ref{sec:mnist}).
A contaminated bandit isolates the mechanism: sign-aware surprisal gating multiplies the perpendicular second moment of contaminated failures by an overlap factor that vanishes as the learner improves, and no gate that depends only on learner probability can match it (Section~\ref{sec:bandit}).
Transformer token reversal then isolates five frictions as controlled dose responses and combines four of them; DG's advantage grows with contamination and sequence length (Section~\ref{sec:results}).

\section{Delightful Policy Gradient}
\label{sec:method}

We briefly recall the Delightful Policy Gradient (DG) of \citet{osband2026delightfulpolicygradient}.
The standard policy gradient forms per-sample updates
$g_t = U_t \nabla_\theta \log \pi_\theta(A_t \mid \mathcal{H}_t),$
where $U_t$ is an advantage estimate and $\mathcal{H}_t$ is the history observed before action $A_t$.
DG augments each term with \emph{action surprisal}
$\ell_t = -\log \pi_\theta(A_t \mid \mathcal{H}_t),$
which measures how unlikely the chosen action was under the learner's current policy.
This surprisal is policy-relative: it depends on the probability the learner assigns to the action, not on how or why the actor generated it.
DG then defines \emph{delight} as their product
$\chi_t = U_t \ell_t$.%
\footnote{We write $\chi$ for the Greek \emph{chara}: delight.}
Delight is positive when an unlikely action has positive advantage and negative when an unlikely action has negative advantage; for actions the policy already expects, surprisal is small and delight stays near zero.

\subsection{Implementation}
\label{sec:implementation}

DG gates each policy-gradient term by
$w_t = \sigma(\chi_t / \eta)$,
where $\sigma(x)=1/(1+e^{-x})$ is the sigmoid and $\eta > 0$ is a temperature.
The resulting update over a batch $\mathcal{B}$, with the gate held fixed when differentiating, is
\begin{equation}
\label{eq:dg}
\Delta \theta \propto \sum_{t \in \mathcal{B}} w_t g_t
= \sum_{t \in \mathcal{B}} \sigma(\chi_t / \eta)\, U_t \,\nabla_\theta \log \pi_\theta(A_t \mid \mathcal{H}_t).
\end{equation}
A breakthrough opens the gate and keeps its update, while a blunder closes the gate and suppresses its update.
For common actions, surprisal is small, so the gate stays near $\tfrac{1}{2}$ and acts as an approximately constant rescaling.
DG is therefore a deliberately biased, learner-relative rule and does not estimate the standard policy gradient.
We use $\eta = 1$ throughout.
DG adds one sigmoid and one multiply per sample, with no measurable wall-clock overhead~\citep{osband2026delightfulpolicygradient}.

REINFORCE weights updates by advantage alone~\citep{williams1992simple}.
Methods based on probability ratios, including importance sampling, PPO-style clipping~\citep{schulman2017proximal}, and V-trace~\citep{espeholt2018impala}, require behavior log-probabilities; DG does not.
\citet{osband2026delightfulpolicygradient} also studies continuous actions using batch-whitened delight to control scale; all experiments here use discrete actions.

Because the gate is nonlinear in the advantage, the scale and baseline of the advantage matter more than in ordinary PG.
Doubling every reward doubles every delight and pushes the gate toward a hard threshold, so advantages should be of order one; all rewards here lie in $[-1,1]$ and we apply no further normalization.
A different baseline can move DG's expected direction, where for PG it changes only the variance.
We use an oracle expected-reward baseline on MNIST, $b{=}1/2$ in the bandit, and the per-prompt mean reward on token reversal; Appendix~\ref{app:mnist_baselines} shows that DG's advantage persists under four baselines, including a zero baseline.

\subsection{Why Delight, Not Importance Weights?}
\label{sec:why}

Like an importance ratio, DG assigns a scalar weight to each policy-gradient term, but the two weights measure different things.
An importance ratio $\pi(a)/\mu(a)$ corrects mismatch between learner and actor, but requires the behavior probability $\mu(a)$, which in distributed systems is often unknown, stale, or corrupted.
Even when available, such ratios are numerically fragile: as the learner improves, it assigns high probability to actions the stale actor rarely took, producing large ratios that force practitioners to clip or truncate and accept the resulting bias~\citep{schulman2017proximal,espeholt2018impala}.

The action was taken and the reward was observed; what matters is not how likely some actor was to generate this sample, but how much the learner can gain from it.
DG therefore asks a different question: how useful is this sample for the learner's current policy?
High delight marks surprising successes that reveal something new; negative delight marks surprising failures on actions the learner has already learned to avoid.
Because surprisal is computed under the learner's current policy, DG remains well-defined even when the actor's policy is unknown.

\section{MNIST Diagnostic}
\label{sec:mnist}

We first show the core limitation in the simplest setting, casting MNIST classification~\citep{lecun1998gradient} as a contextual bandit: given image $x$, the agent samples a label $a \in \{0,\dots,9\}$ from a softmax policy $\pi_\theta$ and receives reward $r = \mathbb{I}\{a = y\}$.
The true label $y$ is never observed, so learning must proceed entirely from reward signal.
We train a two-layer ReLU network with Adam~\citep{kingma2014adam} over batches of $B{=}100$ images, using an oracle expected-reward baseline $b(x)=\pi_\theta(y\mid x)$ for all methods; this baseline uses the true label and is applied identically across methods to remove baseline-quality confounds.
Because the baseline is action-independent, it does not turn the update into supervised learning; the policy still observes only the sampled reward for the chosen action.

We simulate staleness by having the actor use the learner parameters from $D$ gradient steps earlier, modeling a distributed system in which actors lag behind the learner.
We compare three methods.
REINFORCE uses the stale-policy gradient without off-policy correction.
PG applies importance weighting with \emph{exact} behavior probabilities, the strongest possible off-policy correction.
DG ($\eta{=}1$) uses no importance weights at all.
All methods share the same learning rate, batch size, and value baseline; the only difference is how each method weights its gradient update.
Full experimental details, including learning rate selection and baseline sensitivity, appear in Appendix~\ref{app:mnist_details}.

\begin{figure}[ht!]
\centering
\vspace{-1mm}
\begin{subfigure}[t]{0.48\columnwidth}
    \centering
    \includegraphics[width=\linewidth]{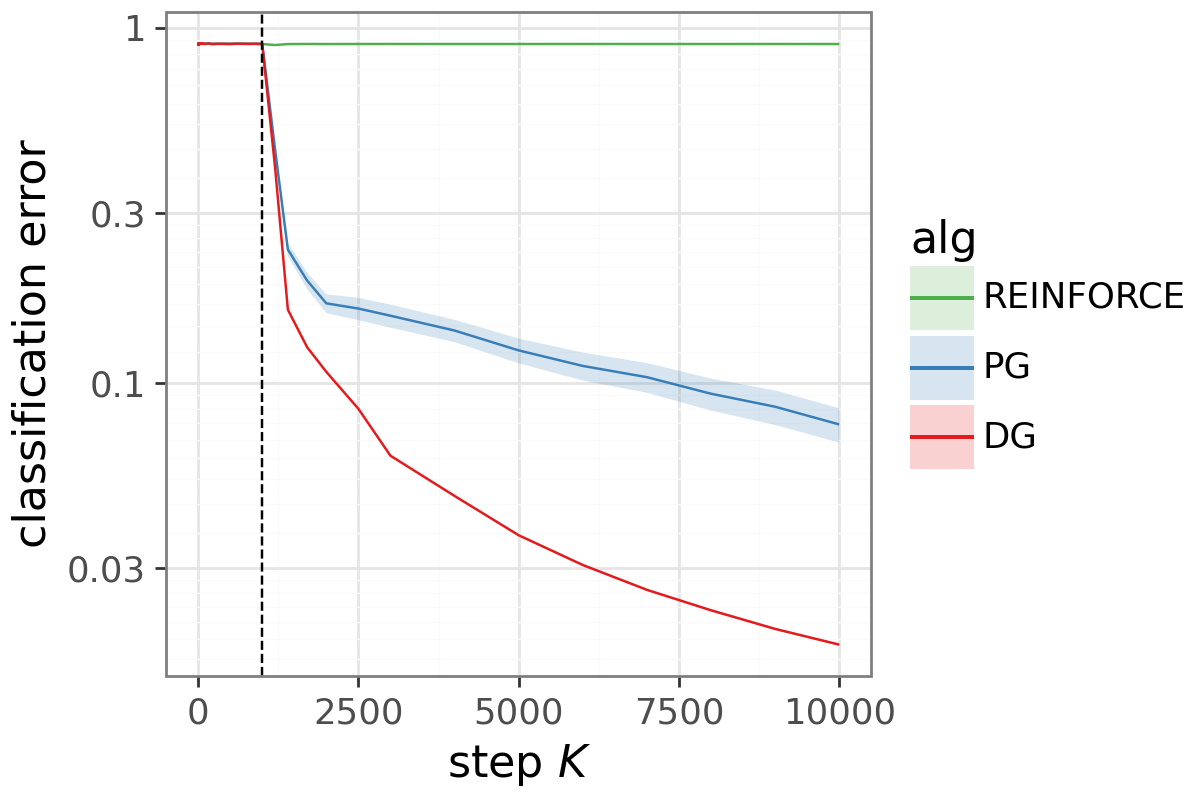}
    \vspace{-3mm}
    \caption{Learning curves at $D{=}1000$.}
    \label{fig:mnist_delay_curve}
\end{subfigure}
\hfill
\begin{subfigure}[t]{0.48\columnwidth}
    \centering
    \includegraphics[width=\linewidth]{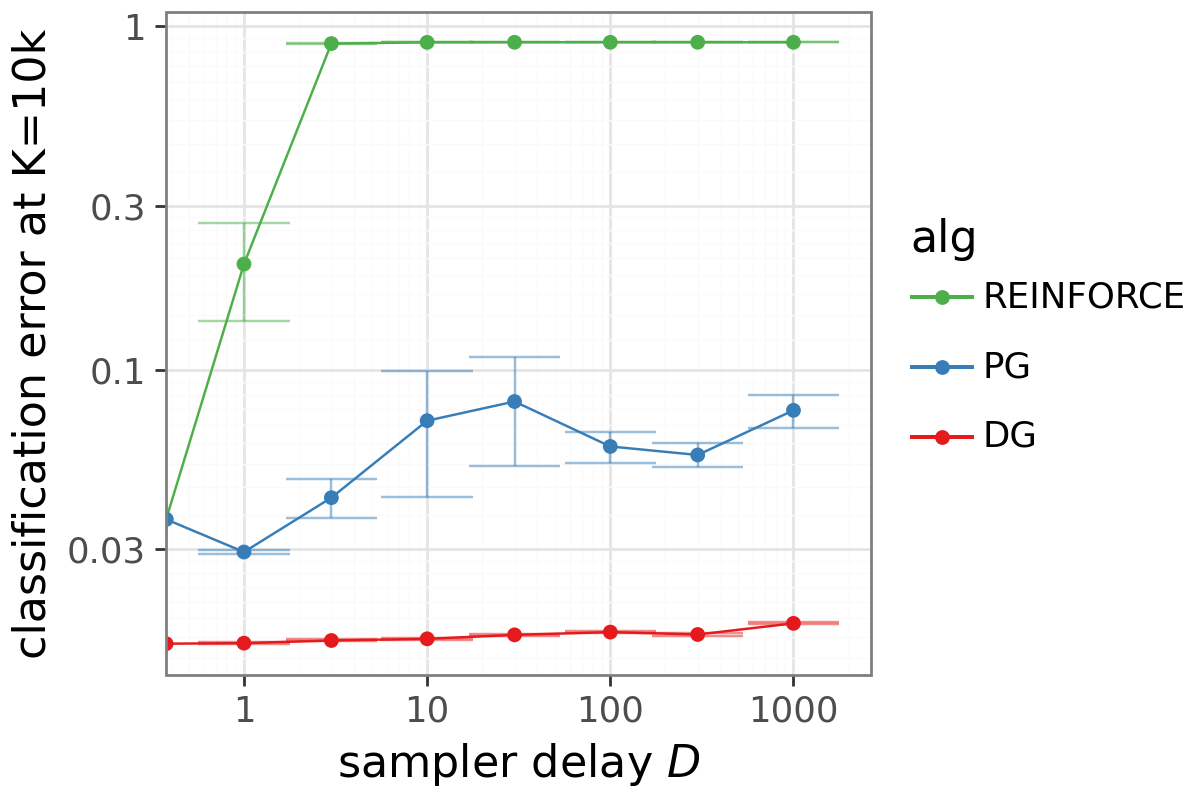}
    \vspace{-3mm}
    \caption{Classification error at $10\text{k}$ gradient steps vs.\ delay $D$.}
    \label{fig:mnist_delay_scale}
\end{subfigure}
\vspace{-1mm}
\caption{MNIST under staleness.
Results average over 30 seeds with $\pm 1$ standard error.}
\label{fig:mnist_delay}
\vspace{-1mm}
\end{figure}

Exact importance weighting only partially repairs the stale gradient, while DG remains strong across the full delay range (Figure~\ref{fig:mnist_delay}).
At $D{=}1000$, REINFORCE fails completely and error remains at $90\%$; the delay sweep shows that performance degrades sharply beyond $D{=}3$ (Figure~\ref{fig:mnist_delay_scale}).
Importance weighting rescues PG from total collapse, but convergence remains slow: after $10\text{k}$ steps, PG reaches roughly $8\%$ error.
DG reaches $2\%$ error over the same horizon, a $4{\times}$ improvement, without using any importance weights.
Across the full delay range, DG stays at or below $2\%$ error, while PG degrades steadily from $3\%$ to $8\%$.

\begin{figure}[ht!]
\centering
\vspace{-1mm}
\begin{subfigure}[t]{0.48\columnwidth}
    \centering
    \includegraphics[width=\linewidth]{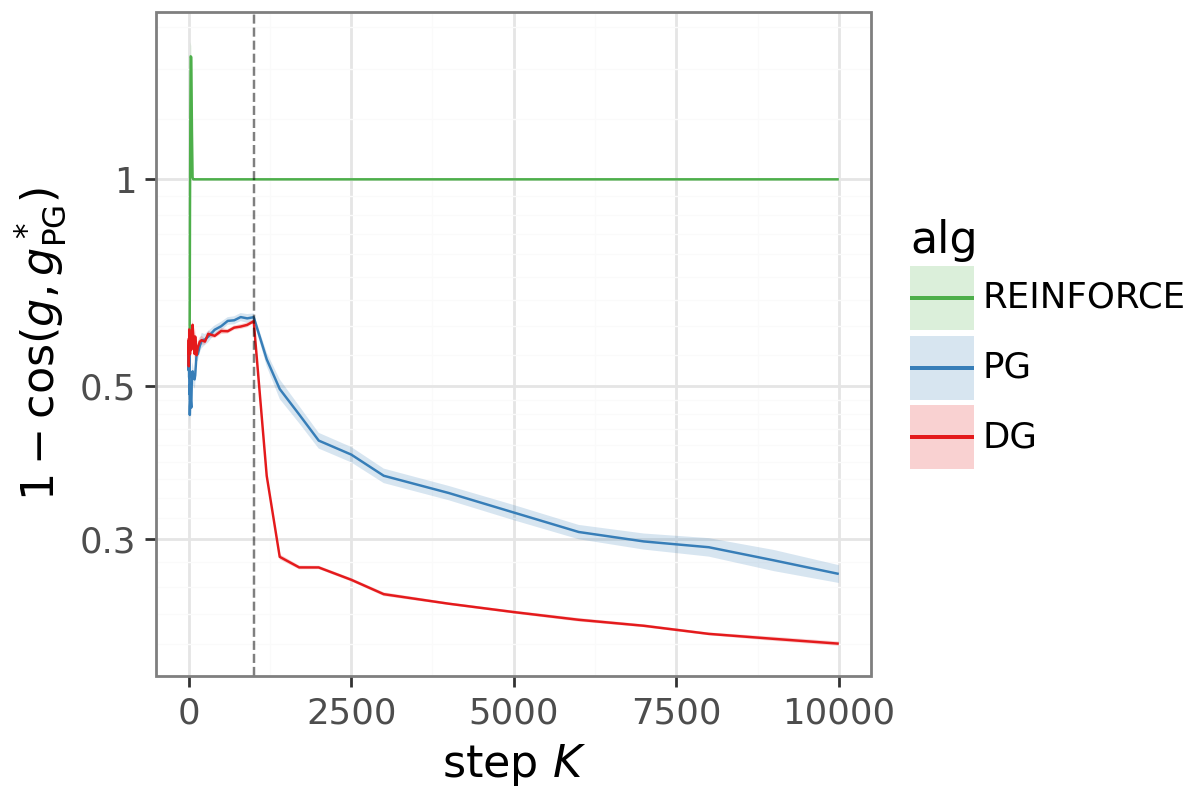}
    \vspace{-3mm}
    \caption{Misalignment versus $g^*_{\mathrm{PG}}$.}
    \label{fig:mnist_grad_pg}
\end{subfigure}
\hfill
\begin{subfigure}[t]{0.48\columnwidth}
    \centering
    \includegraphics[width=\linewidth]{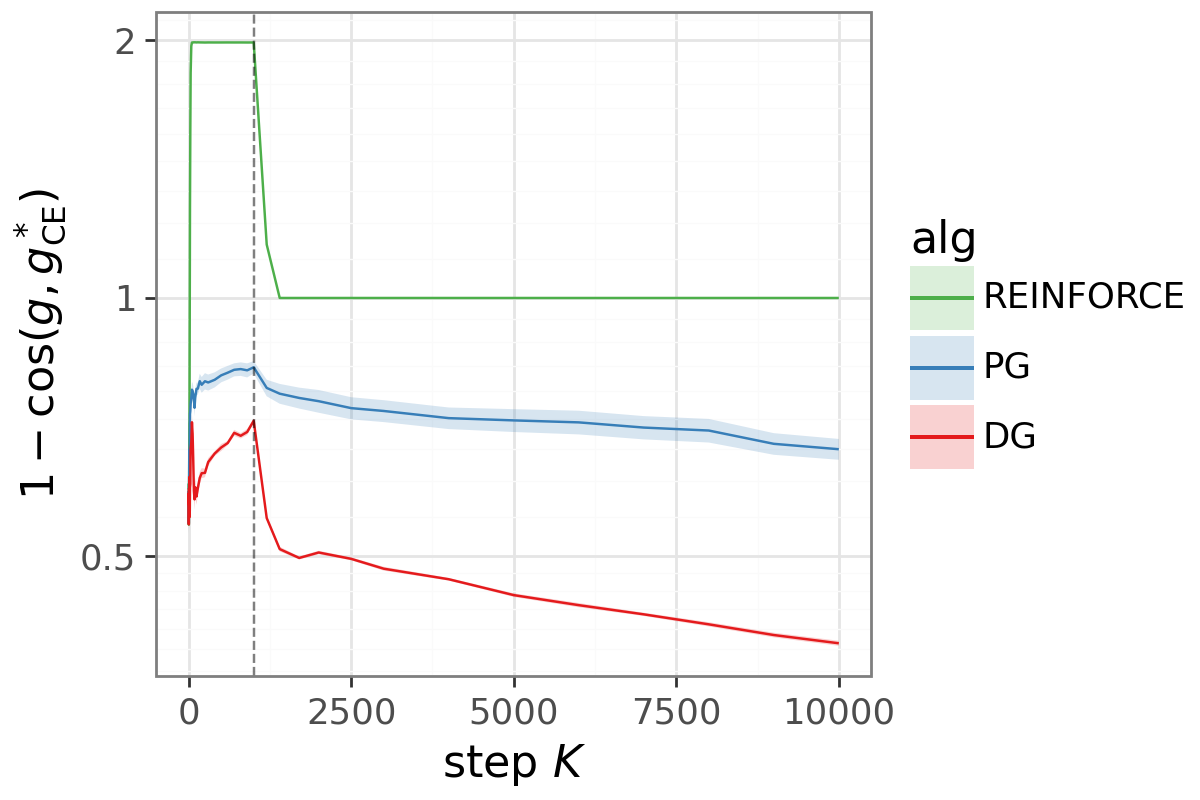}
    \vspace{-3mm}
    \caption{Misalignment versus $g^*_{\mathrm{CE}}$.}
    \label{fig:mnist_grad_ce}
\end{subfigure}
\vspace{-1mm}
\caption{Gradient misalignment under staleness ($D{=}1000$).
\textbf{(a)}~Distance to ideal PG direction $g^*_{\mathrm{PG}}$.
\textbf{(b)}~Distance to cross-entropy direction $g^*_{\mathrm{CE}}$.
The dashed line marks $t{=}D$, one full delay horizon; before it, the actor is still the initial policy.
Results average over 30 seeds with $\pm 1$ standard error.}
\label{fig:mnist_grad}
\vspace{-1mm}
\end{figure}

Figure~\ref{fig:mnist_grad} measures gradient quality directly by plotting $1 - \cos(g, g^*)$ against the ideal policy-gradient direction $g^*_{\mathrm{PG}}$ and the cross-entropy direction $g^*_{\mathrm{CE}}$; lower is better.
REINFORCE gradients are uncorrelated with either target.
PG recovers partial alignment, but DG achieves substantially lower misalignment to both, and the gap grows with training.
Section~\ref{sec:bandit} formalizes one local mechanism behind this pattern in a setting where the relevant gradient geometry is analytically tractable.

This is the best case for off-policy correction, since PG has \emph{exact} behavior probabilities and faces no friction beyond delay, yet DG still outperforms it without importance weights.
The issue is not only how to reconstruct the on-policy gradient under delay, but whether that reconstructed gradient is the best finite-batch learning signal under contamination.
Appendix~\ref{app:mnist_baselines} shows that under a zero baseline, where all advantages are non-negative, REINFORCE tracks DG while PG becomes \emph{worse} than uncorrected REINFORCE, so the mechanism goes beyond positive-only filtering.

\section{Suppressing Rare-Failure Noise in Bandits}
\label{sec:bandit}

We now isolate the mechanism behind DG's robustness in a tabular setting.
The purpose of the analysis is not to claim that every negative update on an incorrect action is useless.
In a single softmax bandit, pushing down an incorrect action can still move probability mass toward the correct action~\citep{osband2026delightfulpolicygradient}.
The problem is more specific: logit score norms $\|\phi_\pi(a)\|$ are bounded, but the uncorrected policy gradient (REINFORCE, as in Section~\ref{sec:mnist}) does not attenuate rare failures as their learner probability vanishes, so contamination allocates finite-batch update mass to score vectors with nonzero perpendicular components.

DG suppresses precisely these high-surprisal negative-advantage terms.
We show two facts.
First, DG suppresses the perpendicular second moment of contaminated failures by a policy-overlap factor that vanishes as the learner assigns those actions low probability.
Second, this requires the sign of the advantage: any learner-probability-only gate that suppresses rare failures also suppresses rare successes.

Consider a $K$-armed bandit with $K\ge3$, correct arm $y^*$, and softmax policy
$\pi=\mathrm{softmax}(z)$ over logits $z\in\mathbb{R}^K$.
Let
\[
\phi_\pi(a) := \nabla_z \log \pi(a) = e_a-\pi,
\qquad
J(z)=\pi(y^*),
\qquad
\nabla_z J = \pi(y^*)\phi_\pi(y^*).
\]
The reward is $r=\mathbb{I}\{a=y^*\}$ with baseline $b=1/2$, so
$U(y^*)=1/2$ and $U(a)=-1/2$ for $a\neq y^*$.
Actions are sampled from a contaminated distribution
$\mu=(1-\rho)\pi+\rho\nu$,
where $\rho$ is the contamination rate and $\nu$ is an arbitrary distribution over actions.
Gradients are computed under the current learner policy $\pi$.
Let $\ell(a):=-\log\pi(a)$ denote surprisal and write $w(a):=\sigma\!\big(U(a)\ell(a)/\eta\big)$ for the DG gate.
Write $u := \phi_\pi(y^*)/\|\phi_\pi(y^*)\|$ for the unit true-gradient direction and $\Pi_\perp := I - uu^\top$ for orthogonal projection.

\begin{lemma}[Asymmetric tail gating]
\label{lem:tail_gate}
Let an action have probability $p$ under the learner and advantage magnitude $c>0$.
For positive advantage,
$\sigma(c(-\log p)/\eta) \ge 1-p^{c/\eta}$.
For negative advantage,
$\sigma(-c(-\log p)/\eta) \le p^{c/\eta}$.
Thus, as $p\to 0$, DG preserves rare successes and suppresses rare failures.
\end{lemma}

For the bandit above, $c=1/2$, so each disfavored failure receives gate at most
$\pi(a)^{1/(2\eta)}$.

\begin{proposition}[Perpendicular failure noise is suppressed]
\label{prop:perp_suppression}
Let $G_{\mathrm{RF}}(a)=U(a)\phi_\pi(a)$ and $G_{\mathrm{DG}}(a)=w(a)U(a)\phi_\pi(a)$.
Define the failure-only perpendicular second moment
$
V^{\mathrm{alg}}_{\perp,\mathrm{fail}}
:=
\mathbb{E}_{a\sim\mu}
\big[
\|\Pi_\perp G_{\mathrm{alg}}(a)\|^2
\,\mathbb{I}\{a\neq y^*\}
\big].
$
Then $V^{\mathrm{RF}}_{\perp,\mathrm{fail}} = \tfrac{1}{4} \sum_{a\neq y^*} \mu(a) \|\Pi_\perp\phi_\pi(a)\|^2$,
while
$V^{\mathrm{DG}}_{\perp,\mathrm{fail}} \le \tfrac{1}{4} \sum_{a\neq y^*} \mu(a)\,\pi(a)^{1/\eta} \|\Pi_\perp\phi_\pi(a)\|^2$.
\end{proposition}

The contaminated contribution is controlled by the overlap moment
$M_{\nu,\eta}^{\perp}(\pi) := \sum_{a\neq y^*} \nu(a)\pi(a)^{1/\eta} \|\Pi_\perp\phi_\pi(a)\|^2$.
As the learner improves and $\max_{a\neq y^*}\pi(a)\to0$, this overlap moment vanishes for every fixed $\eta>0$.
For the default $\eta=1$, $M_{\nu,1}^{\perp}(\pi) \le 2\max_{a\neq y^*}\pi(a) \to 0$, using $\|\phi_\pi(a)\|^2 = \|e_a - \pi\|^2 \le 2$.
By contrast, REINFORCE's contaminated perpendicular second moment contains the term
\[
\frac{\rho}{4}\sum_{a\neq y^*}\nu(a)\|\Pi_\perp\phi_\pi(a)\|^2 .
\]
It is $\Omega(\rho)$ whenever there exists a constant $\gamma>0$ such that
\[
\sum_{a\neq y^*}\nu(a)\|\Pi_\perp\phi_\pi(a)\|^2 \ge \gamma
\]
throughout the near-optimal regime.
This condition holds whenever $\nu$ places mass on at least two distinct incorrect actions, as uniform contamination does.

As the learner improves, disfavored contaminated actions become lower probability under the learner; DG then suppresses their perpendicular contribution more strongly, reducing the noise that can rotate the minibatch update.
Because the gate conditions on the advantage sign, it suppresses learner-rare failures while keeping learner-rare successes.
Proposition~\ref{prop:sign_blind_tradeoff} shows that this suppression requires the advantage sign.

\begin{proposition}[Learner-probability-only gates cannot separate rare successes from rare failures]
\label{prop:sign_blind_tradeoff}
Let $q(p)\in[0,\infty)$ be any gate that depends on an action only through its learner probability $p=\pi(a)$ and not on the advantage sign.
If $q(p)\to0$ as $p\to0$, then the gate suppresses both rare failures and rare successes.
If $\liminf_{p\to0}q(p)>0$, then it does not suppress rare failures.
DG avoids this tradeoff: for any fixed advantage magnitude $c>0$,
$w_+(p):=\sigma(c(-\log p)/\eta)\to1$ while $w_-(p):=\sigma(-c(-\log p)/\eta)\to0$.
\end{proposition}

In a single softmax bandit near the optimum, the absolute cosine between an incorrect-action score vector $\phi_\pi(a)$ and $\nabla_z J$ is $\Theta(1)$; equivalently, the negative update $-\phi_\pi(a)$ has positive $\Theta(1)$ cosine with the true-gradient direction.
So the population direction need not collapse; DG instead suppresses the perpendicular second moment of rare failures, increasingly so as the learner improves.
In sequential and multi-context problems, contaminated failures often correspond to stale trajectories, actor bugs, or misleading reward assignments.
Perpendicular suppression prevents finite-batch update budget from being dominated by such high-surprisal negative samples.
Exact importance weighting $f(a)=\pi(a)/\mu(a)$ removes contamination bias at the population level by reconstructing the on-policy expected gradient (Remark~\ref{rem:exact_is}), but requires accurate behavior probabilities.
Proofs appear in Appendix~\ref{app:bandit_proofs}.

Figures~\ref{fig:bandit_dynamics} and~\ref{fig:bandit_scaling} show the corresponding empirical pattern: the rare-failure suppression mechanism translates into higher cosine similarity during training and lower suboptimality across contamination levels.

\begin{figure}[ht!]
\centering
\vspace{-1mm}
\begin{subfigure}[t]{0.48\columnwidth}
    \centering
    \includegraphics[width=\linewidth]{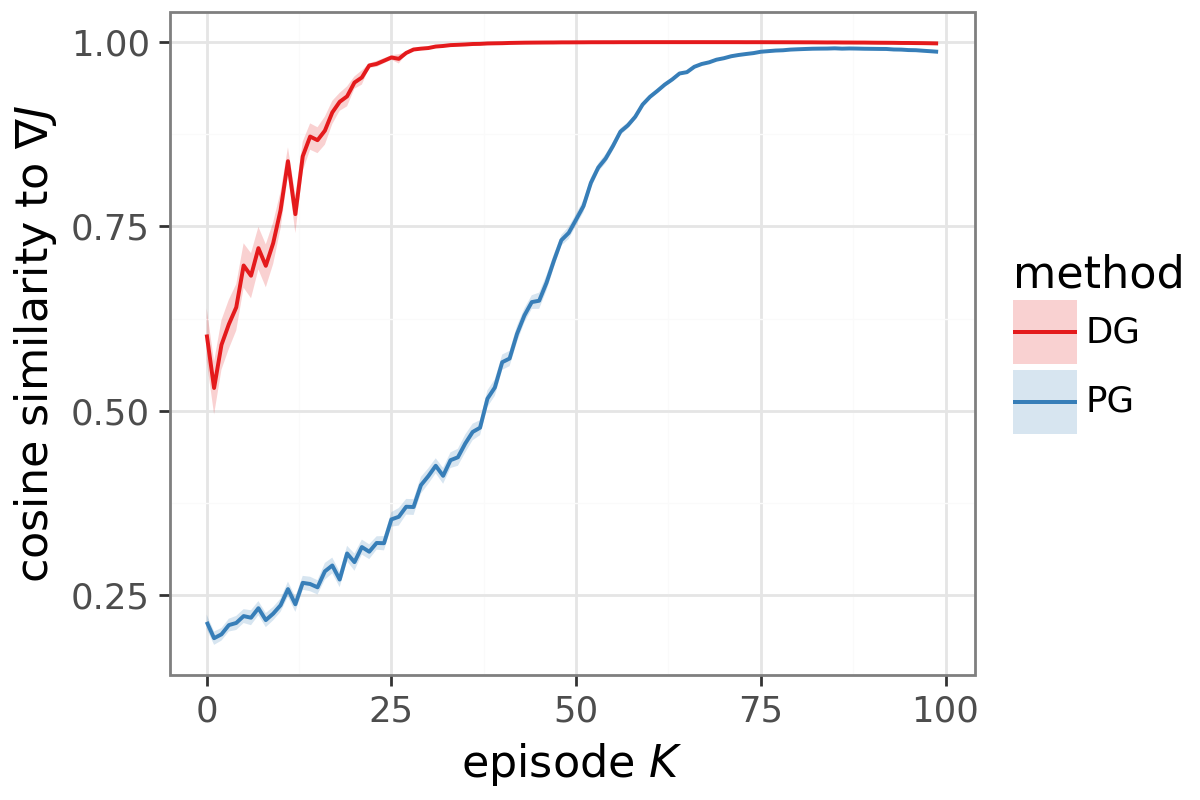}
    \vspace{-3mm}
    \caption{Cosine similarity to $\nabla_z J$ during training ($\rho{=}0.1$).}
    \label{fig:bandit_dynamics}
\end{subfigure}
\hfill
\begin{subfigure}[t]{0.48\columnwidth}
    \centering
    \includegraphics[width=\linewidth]{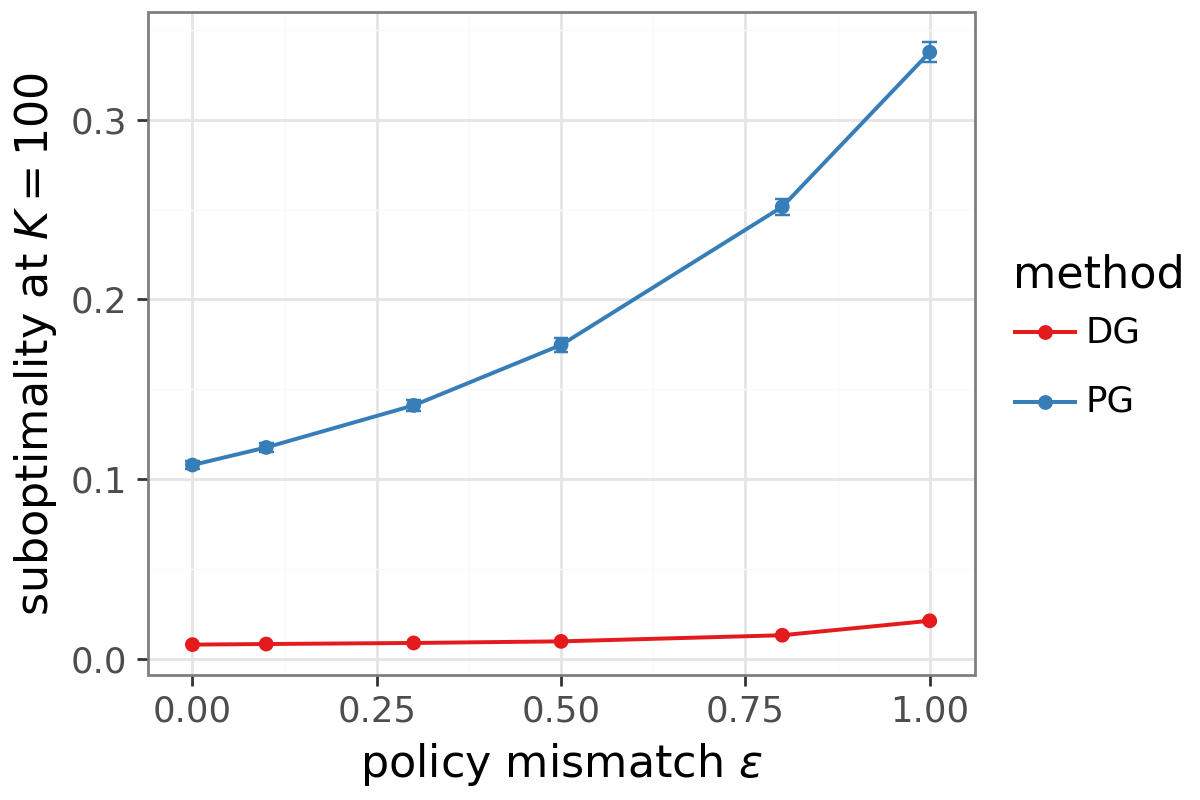}
    \vspace{-3mm}
    \caption{Suboptimality $1{-}\pi(y^*)$ vs.\ contamination $\rho$.}
    \label{fig:bandit_scaling}
\end{subfigure}
\vspace{-1mm}
\caption{$K$-armed bandit ($K{=}100$), $\nu=\mathrm{Unif}([K])$, $B{=}100$, $\alpha{=}0.1$, $\eta{=}1$.
DG suppresses rare-failure perpendicular noise, maintaining gradient alignment and lower suboptimality.
Results average over 100 seeds with $\pm 1$ standard error.}
\label{fig:bandit}
\vspace{-1mm}
\end{figure}

We next test whether the same advantage survives sequential decisions, function approximation, and multiple frictions at once.

\section{Token Reversal with Distributed Friction}
\label{sec:results}

The bandit analysis isolates a single contaminated decision with exact gradients.
We now test whether the same local selectivity survives in token reversal~\citep{osband2026delightfulpolicygradient}, a transformer sequence task that captures the core pattern of reasoning-style language-model training: reading an input, preserving structure, and generating a coherent output autoregressively.
As sequence length grows, correct trajectories become exponentially rarer, so distributed frictions become more consequential.

\begin{figure}[ht!]
\centering
\includegraphics[width=0.6\columnwidth]{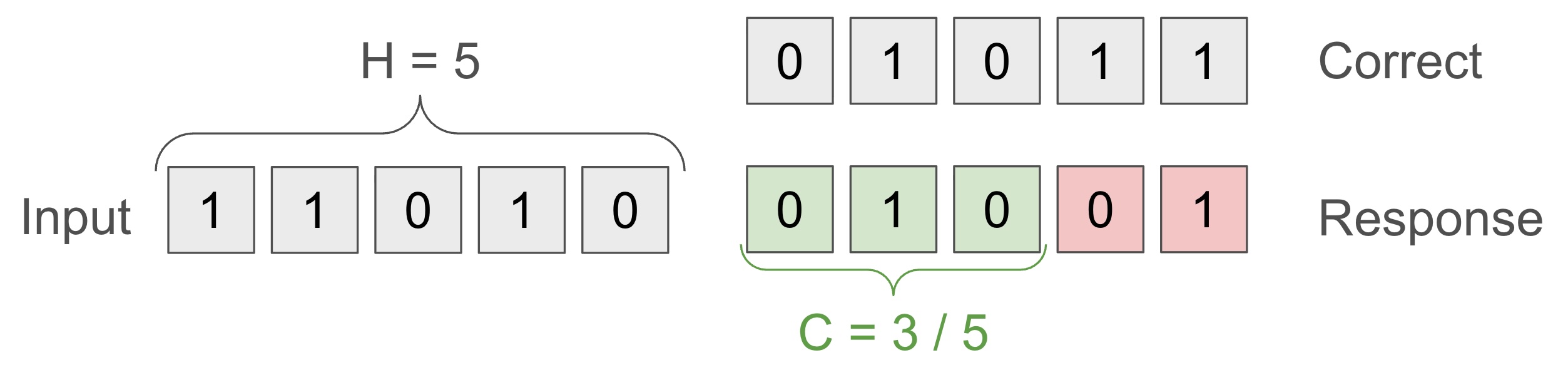}
\vspace{-2mm}
\caption{Token reversal ($M{=}2$, $H{=}5$): the agent must output the input in reverse. Here it gets three correct then errs, giving $c_k = 3/5$ and reward $R_k = \kappa \cdot 3/5$.}
\label{fig:binary_reverse_strict}
\end{figure}

The agent receives $H$ tokens drawn uniformly from a vocabulary of size $M$ and must output them in reverse order (Figure~\ref{fig:binary_reverse_strict}).
Let $N_k$ denote the number of consecutive correct tokens from the start of episode $k$, and define correctness $c_k = N_k / H$.
We extend the task with a reward-shaping parameter $\kappa \in [-1, 1]$: $R_k = \kappa \, c_k + (1 - \kappa) \, \mathbb{I}\{c_k = 1\}$.
When $\kappa > 0$ (\emph{hedonic guide}), partial progress is rewarded; when $\kappa < 0$ (\emph{hedonic trap}), it is penalized, modeling settings where easy shortcuts do not generalize to full solutions.
We compare DG against PG (importance-weighted policy gradient), PPO~\citep{schulman2017proximal}, and PMPO~\citep{abdolmaleki2024preference}.
Unless otherwise noted, we use $M{=}2$, $H{=}10$, $\kappa{=}1$, and $K{=}1000$ gradient steps with $100$ episodes per step ($10$ prompts ${\times}$ $10$ responses; $K$ counts gradient steps in all plots), over 30 seeds; details appear in Appendix~\ref{app:token_reversal_details}.
We report sequence error $1 - \mathbb{E}[R_k]$ on the trajectories the actor generates; under the default $\kappa{=}1$ this equals $1 - \mathbb{E}[c_k]$.
All methods share the GRPO-style grouped baseline, the mean reward of the 10 responses to each prompt~\citep{shao2024deepseekmath}, so our PPO is GRPO without its standard-deviation normalization and KL penalty.
Code is available at \codeurl.

The frictions differ in how directly distributed execution causes them.
Staleness and probability mismatch arise from distributed execution itself: asynchronous actors run on weights that lag the learner, the defining trade-off of IMPALA, Ape-X, and SEED~\citep{espeholt2018impala,horgan2018distributed,espeholt2020seed}, and even at zero lag separate inference and training stacks can assign different probabilities to the same token~\citep{he2025nondeterminism,yao2025offpolicy}.
Actor bugs and reward corruption can occur in any pipeline, although large distributed pipelines make them more likely: a truncated generation or corrupted rollout file produces a trajectory no recent learner would emit, and policy-gradient results are sensitive to such implementation details~\citep{engstrom2020implementation}; reward models and verifiers make errors that a policy can exploit~\citep{everitt2017corrupted,gao2023scaling,skalse2022defining}.
Rare discovery tests the opposite case, a surprising success: clipping and importance correction suppress rare successes along with rare failures, so this test asks whether a method can reject blunders and still learn from a breakthrough, as when correct reasoning rollouts are initially rare.

Under each friction, DG outperforms every tuned baseline, often by an order of magnitude.
We then combine staleness, actor bugs, reward corruption, and rare discovery, tune once at $H{=}5$, and scale sequence length.

\subsection{Staleness}
\label{sec:staleness}

Staleness creates high-surprisal failures from old policies.
We model this by having each actor use a policy sampled uniformly from a buffer of the last $D{+}1$ learner checkpoints, initialized with the starting parameters.
Every point uses the same $K{=}1000$ learner updates, and sequence error measures the delayed actor, the policy that actually generates trajectories.
At delay $D{=}30$, many logged actions have high surprisal under the current learner because the old policy favored actions the learner has since learned to avoid.

\begin{figure}[ht!]
\centering
\vspace{-1mm}
\begin{subfigure}[t]{0.48\columnwidth}
    \centering
    \includegraphics[width=\linewidth]{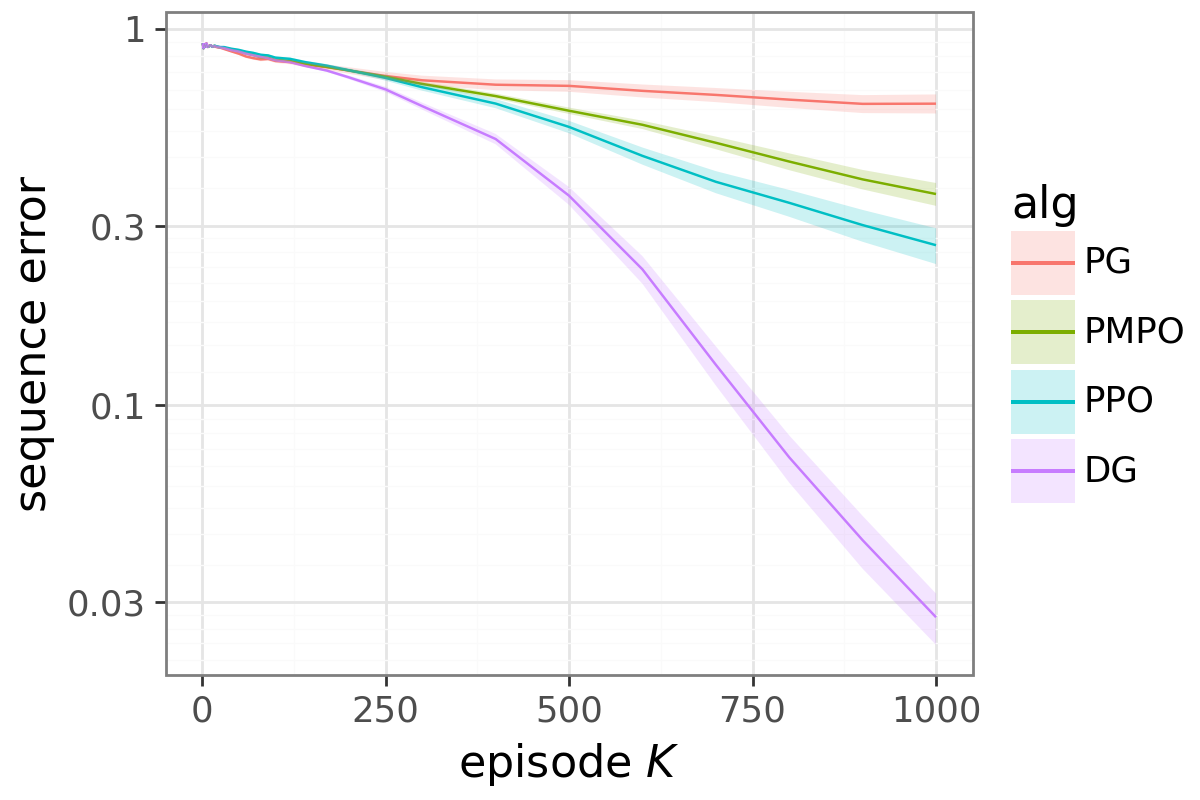}
    \vspace{-2mm}
    \caption{Learning curves at $D{=}30$.}
    \label{fig:delay_regret}
\end{subfigure}
\hfill
\begin{subfigure}[t]{0.48\columnwidth}
    \centering
    \includegraphics[width=\linewidth]{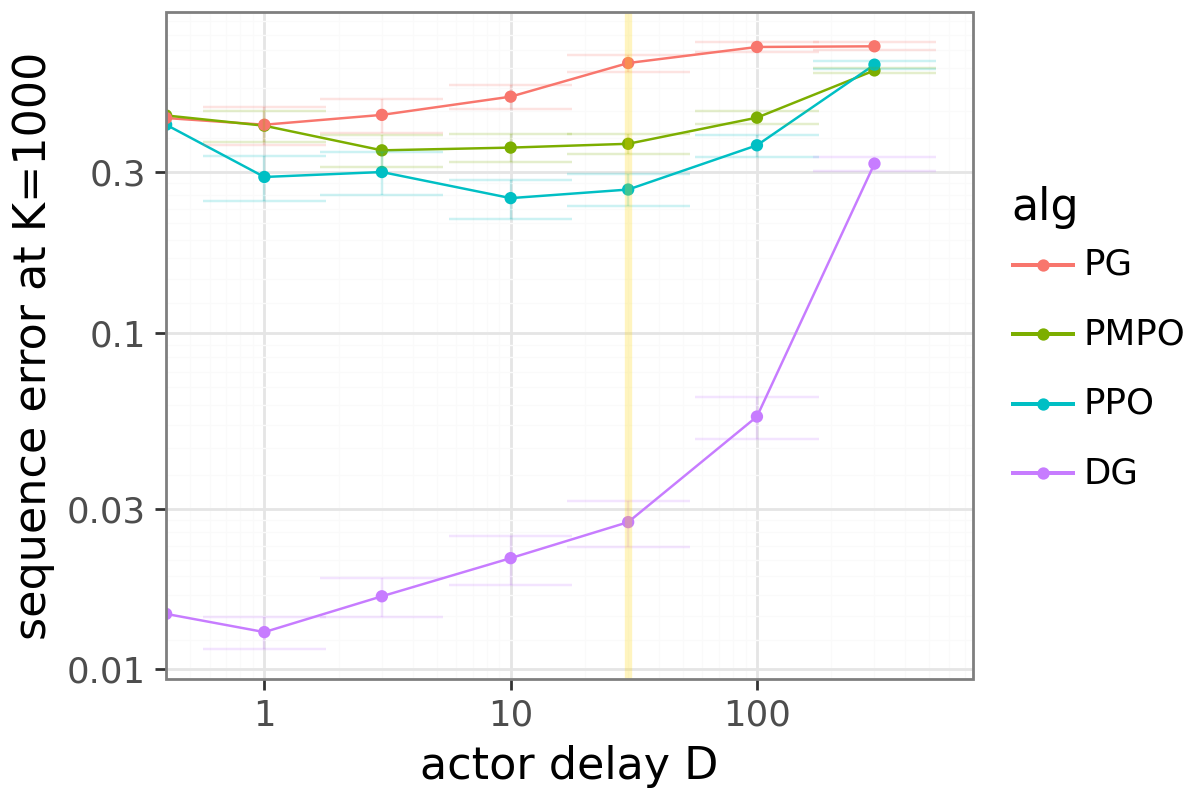}
    \vspace{-2mm}
    \caption{Sequence error at $K{=}1000$ vs.\ delay $D$.}
    \label{fig:delay_scaling}
\end{subfigure}
\vspace{-1mm}
\caption{Sensitivity to actor delay, evaluated on the delayed actor. All methods tuned at $D{=}30$. DG outperforms every baseline across the full range; even with large delay it beats baselines at $D{=}1$.}
\vspace{-1mm}
\label{fig:delay}
\end{figure}

Under staleness, DG converges to near-zero error where baselines stall an order of magnitude higher (Figure~\ref{fig:delay}).
The held-out sweep shows the same pattern across delays: even at $D{=}100$, DG outperforms every baseline at $D{=}1$.
Staleness that cripples PG and PPO degrades DG only gradually, because delight is computed under the learner's current policy rather than the actor's.
We show $D \le 300$: as $D$ approaches $K$, the buffer still holds early checkpoints at the end of training, so actor performance says little about any method.

\FloatBarrier

\subsection{Probability Mismatch}
\label{sec:mismatch}

Probability mismatch creates surprising actions even when actor and learner share weights.
We model this by having the actor sample from $\mathrm{softmax}(z+\epsilon)$ while the learner scores the clean logits $z$.
The perturbation $\epsilon \sim \mathcal{N}(0,\sigma^2 I_M)$ is drawn once per run and added at every position, a persistent bias of the kind a systematic mismatch between inference and training stacks would produce.
PG, PPO, and PMPO receive the exact perturbed behavior probabilities, the best case for ratio-based correction; DG uses none.

\begin{figure}[ht!]
\centering
\vspace{-1mm}
\begin{subfigure}[t]{0.48\columnwidth}
    \centering
    \includegraphics[width=\linewidth]{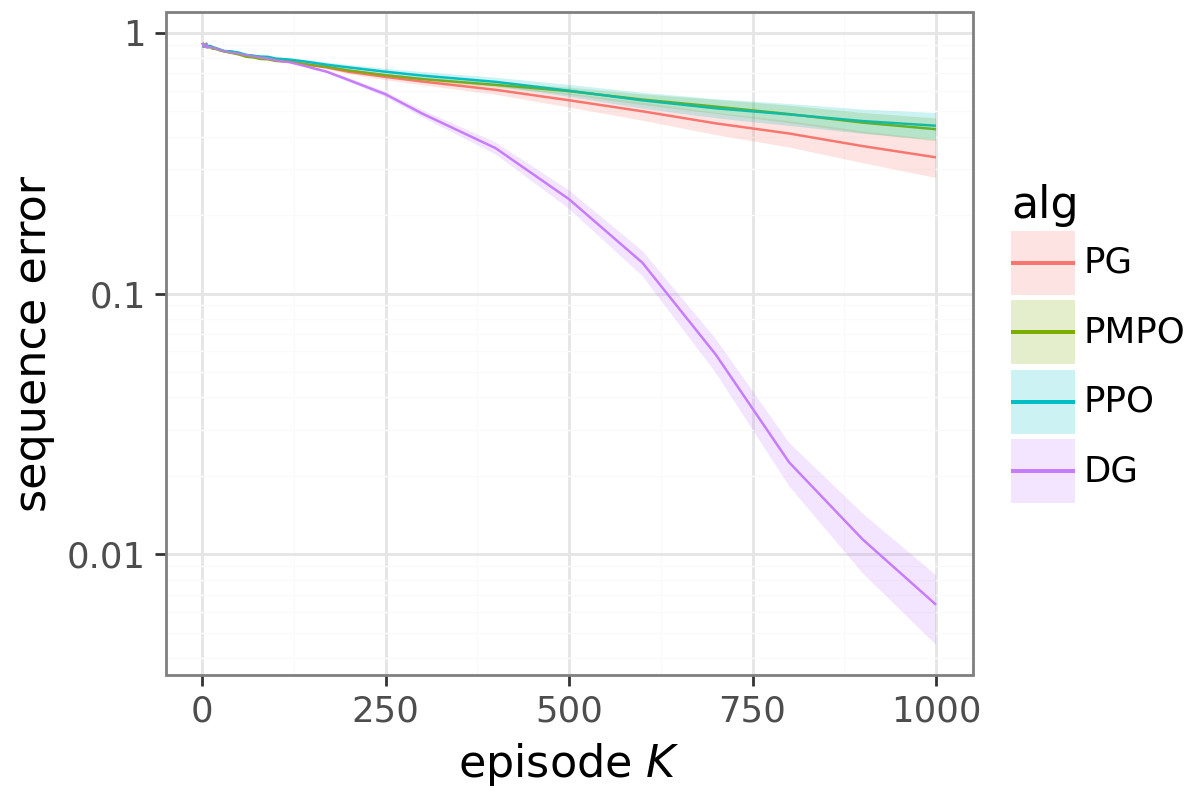}
    \vspace{-2mm}
    \caption{Learning curves at $\sigma{=}0.3$.}
    \label{fig:mismatch_regret}
\end{subfigure}
\hfill
\begin{subfigure}[t]{0.48\columnwidth}
    \centering
    \includegraphics[width=\linewidth]{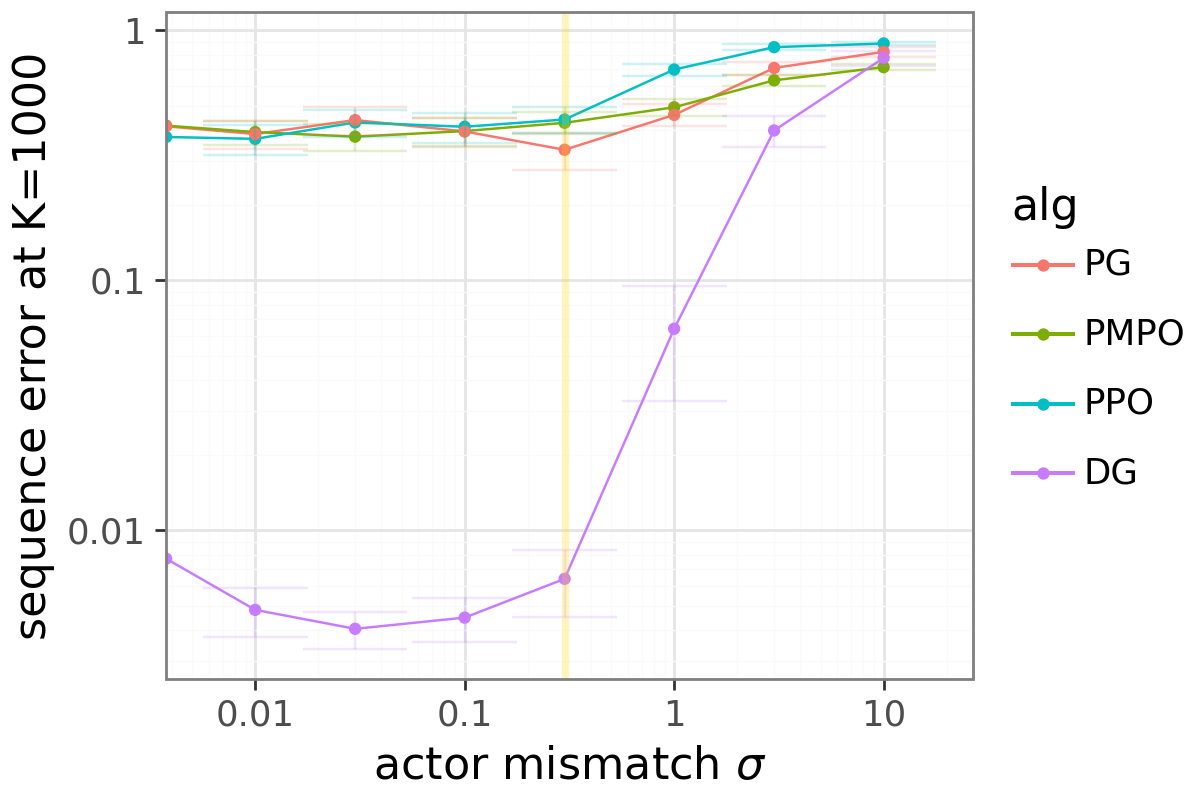}
    \vspace{-2mm}
    \caption{Sequence error at $K{=}1000$ vs.\ logit noise $\sigma$.}
    \label{fig:mismatch_scaling}
\end{subfigure}
\vspace{-1mm}
\caption{Sensitivity to actor--learner probability mismatch under identical weights. All methods tuned at $\sigma{=}0.3$; baselines receive exact behavior probabilities.}
\vspace{-1mm}
\label{fig:mismatch}
\end{figure}

Figure~\ref{fig:mismatch} shows that at $\sigma{=}0.3$, DG drives sequence error below $1\%$ while every baseline stays above $30\%$.
Small $\sigma$ produces the low-surprisal per-token drift typical of numerical mismatch, and DG is unaffected by it; the baselines' error there is their friction-free level at this budget.
The sweep shows the same gap for every $\sigma \le 0.3$, and at $\sigma{=}1$ DG still has roughly $7{\times}$ lower error than the best baseline.
At $\sigma \ge 3$, where the actor's samples bear little relation to the learner's policy, all methods fail.
This controlled same-weights mismatch does not reproduce any particular inference backend.

\FloatBarrier

\subsection{Actor Bugs}
\label{sec:bugs}

Actor bugs create trajectories that are highly surprising and almost always wrong.
We model this by forcing an actor to emit an all-zeros trajectory with probability $p_E$.
These trajectories have high surprisal and negative advantage, so DG should close the gate and suppress them.

\begin{figure}[ht!]
\centering
\vspace{-1mm}
\begin{subfigure}[t]{0.48\columnwidth}
    \centering
    \includegraphics[width=\linewidth]{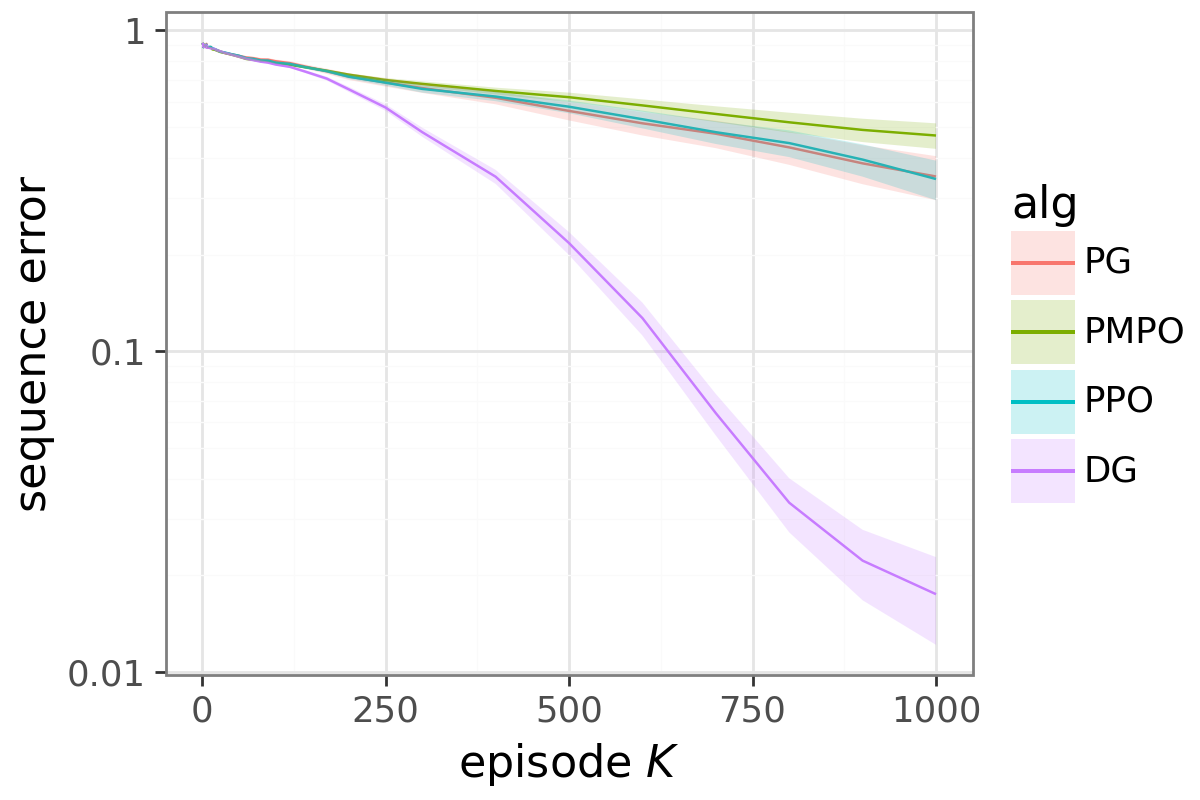}
    \vspace{-2mm}
    \caption{Learning curves at $p_E{=}3{\times}10^{-3}$.}
    \label{fig:actor_bug_regret}
\end{subfigure}
\hfill
\begin{subfigure}[t]{0.48\columnwidth}
    \centering
    \includegraphics[width=\linewidth]{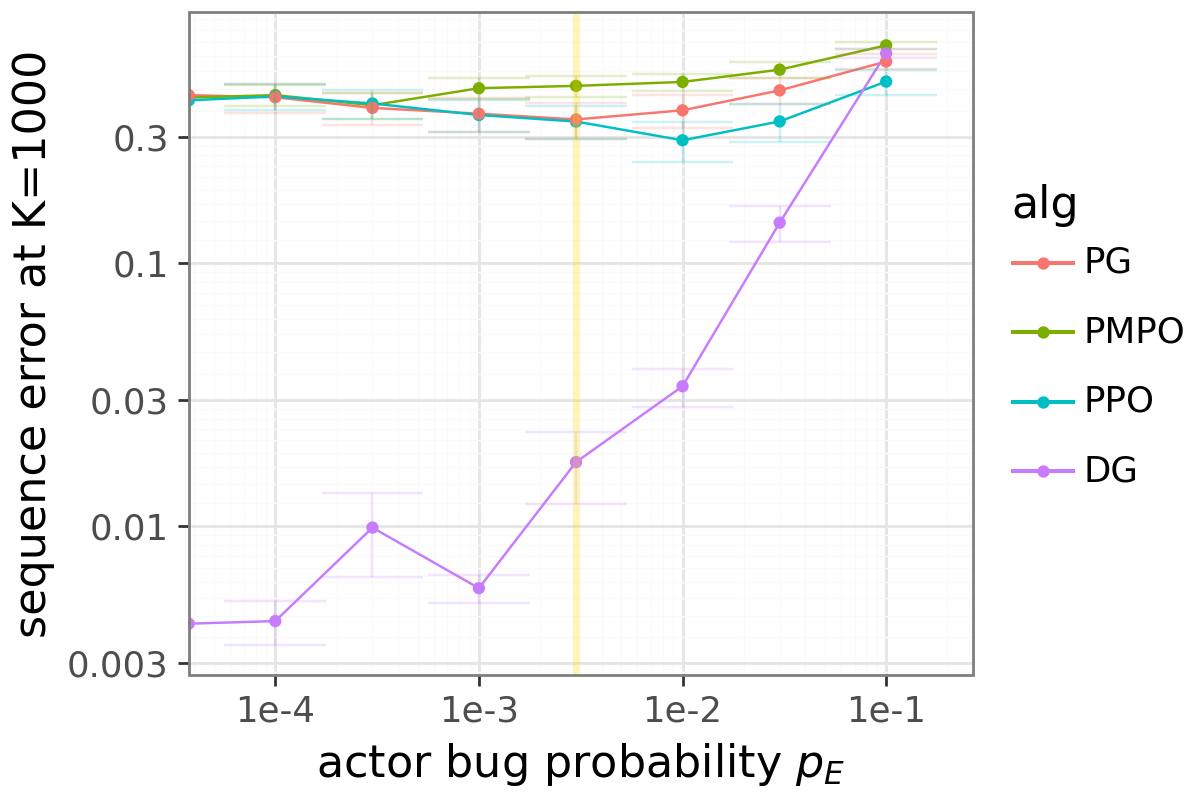}
    \vspace{-2mm}
    \caption{Sequence error at $K{=}1000$ vs.\ bug probability $p_E$.}
    \label{fig:actor_bug_scaling}
\end{subfigure}
\vspace{-1mm}
\caption{Sensitivity to actor bugs. All methods tuned at $p_E{=}3{\times}10^{-3}$. DG outperforms every baseline across the full range of bug probabilities.}
\vspace{-1mm}
\label{fig:actor_bug}
\end{figure}

At $p_E{=}3{\times}10^{-3}$ (Figure~\ref{fig:actor_bug}), DG converges to low error while baselines plateau above $10\%$.
The sweep shows robustness over several orders of magnitude: even at $p_E{=}10^{-2}$, DG outperforms every baseline run without bugs.

\FloatBarrier

\subsection{Reward Corruption}
\label{sec:reward}

Reward corruption creates misleading advantage estimates even when the trajectory itself is unremarkable.
We model this by replacing the episode reward with an independent $\mathrm{Bernoulli}(0.5)$ draw with probability $p_R$.
Unlike staleness and bugs, this friction corrupts the advantage rather than the action distribution.
This tests robustness to errors in the advantage, which the bandit analysis does not cover.

\begin{figure}[ht!]
\centering
\vspace{-1mm}
\begin{subfigure}[t]{0.48\columnwidth}
    \centering
    \includegraphics[width=\linewidth]{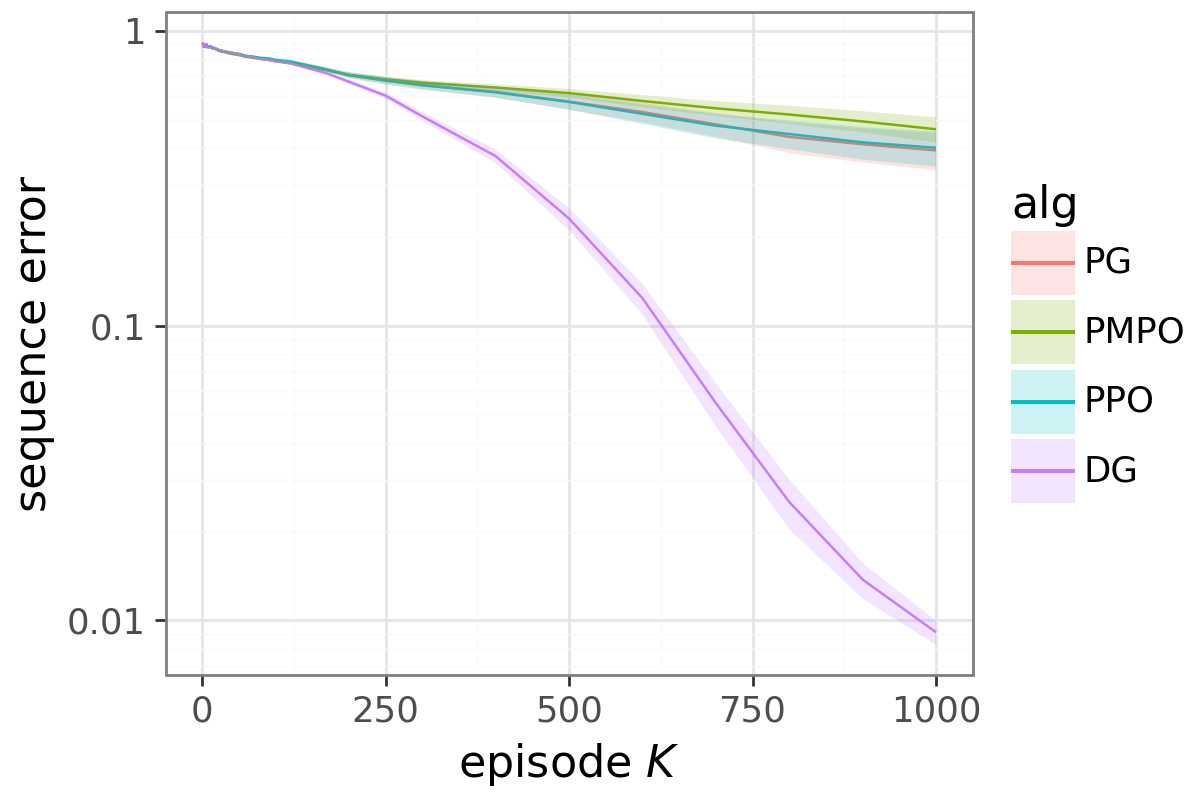}
    \vspace{-2mm}
    \caption{Learning curves at $p_R{=}0.01$.}
    \label{fig:noise_regret}
\end{subfigure}
\hfill
\begin{subfigure}[t]{0.48\columnwidth}
    \centering
    \includegraphics[width=\linewidth]{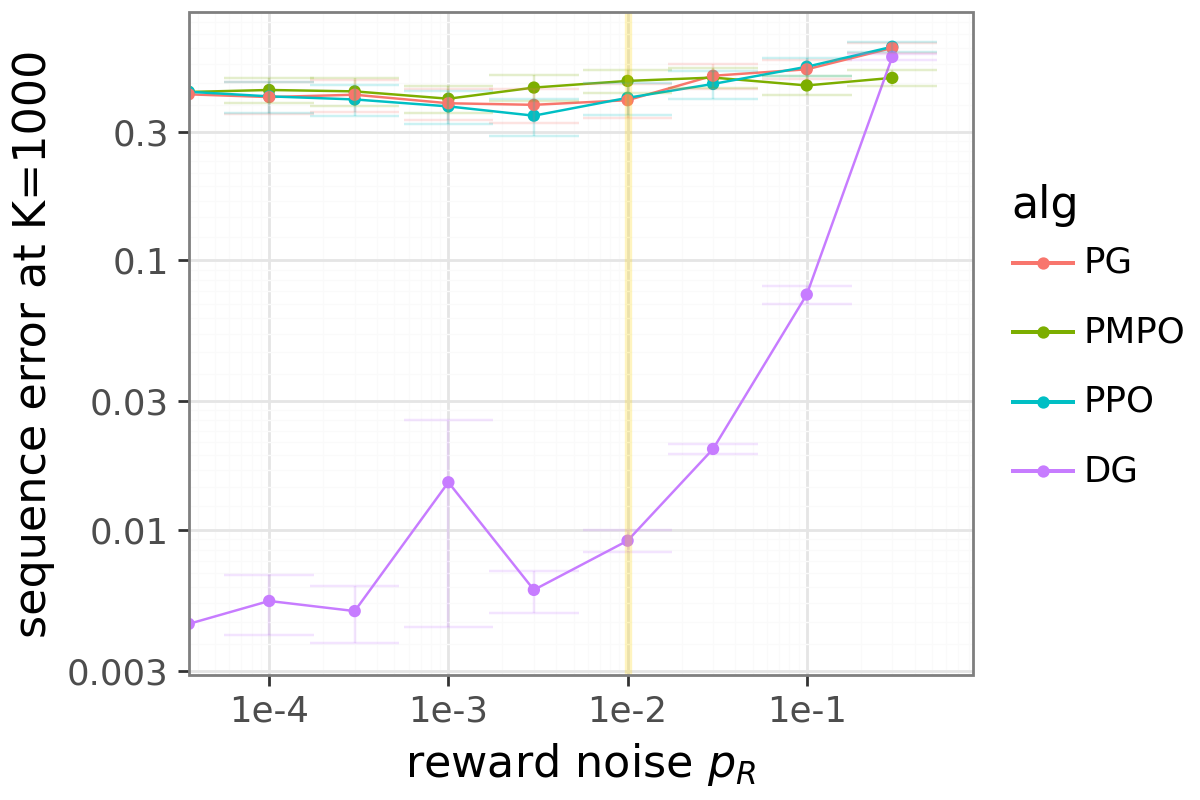}
    \vspace{-2mm}
    \caption{Sequence error at $K{=}1000$ vs.\ reward noise $p_R$.}
    \label{fig:noise_scaling}
\end{subfigure}
\vspace{-1mm}
\caption{Sensitivity to reward corruption. All methods tuned at $p_R{=}0.01$. DG outperforms every baseline across the full range of corruption rates.}
\vspace{-1mm}
\label{fig:noise}
\end{figure}

At $p_R{=}0.01$, DG reaches substantially lower error than every baseline (Figure~\ref{fig:noise}).
The sweep confirms the same pattern: DG maintains low error for $p_R \lesssim 10^{-2}$, while baselines remain at high error across the full range.

\FloatBarrier

\subsection{Rare Discovery}
\label{sec:discovery}

Rare discovery tests the rare-success preservation side of the gate.
We switch to the hedonic trap ($\kappa{=}{-}1$, overriding the default $\kappa{=}1$) with $H{=}5$, where only perfect trajectories yield positive reward, and inject oracle episodes with probability $p_C$.
These episodes are both surprising and successful, so they produce large positive delight.

\begin{figure}[ht!]
\centering
\vspace{-1mm}
\begin{subfigure}[t]{0.48\columnwidth}
    \centering
    \includegraphics[width=\linewidth]{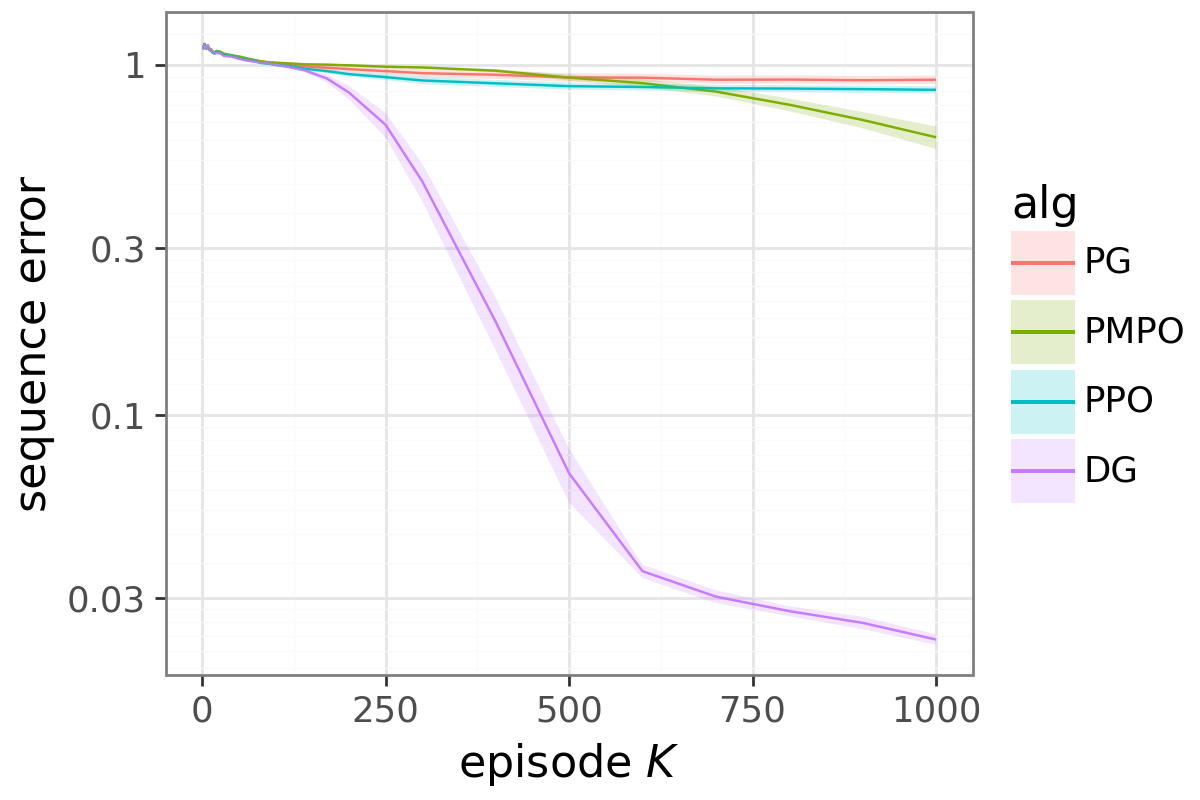}
    \vspace{-2mm}
    \caption{Learning curves at $p_C{=}10^{-3}$.}
    \label{fig:discovery_regret}
\end{subfigure}
\hfill
\begin{subfigure}[t]{0.48\columnwidth}
    \centering
    \includegraphics[width=\linewidth]{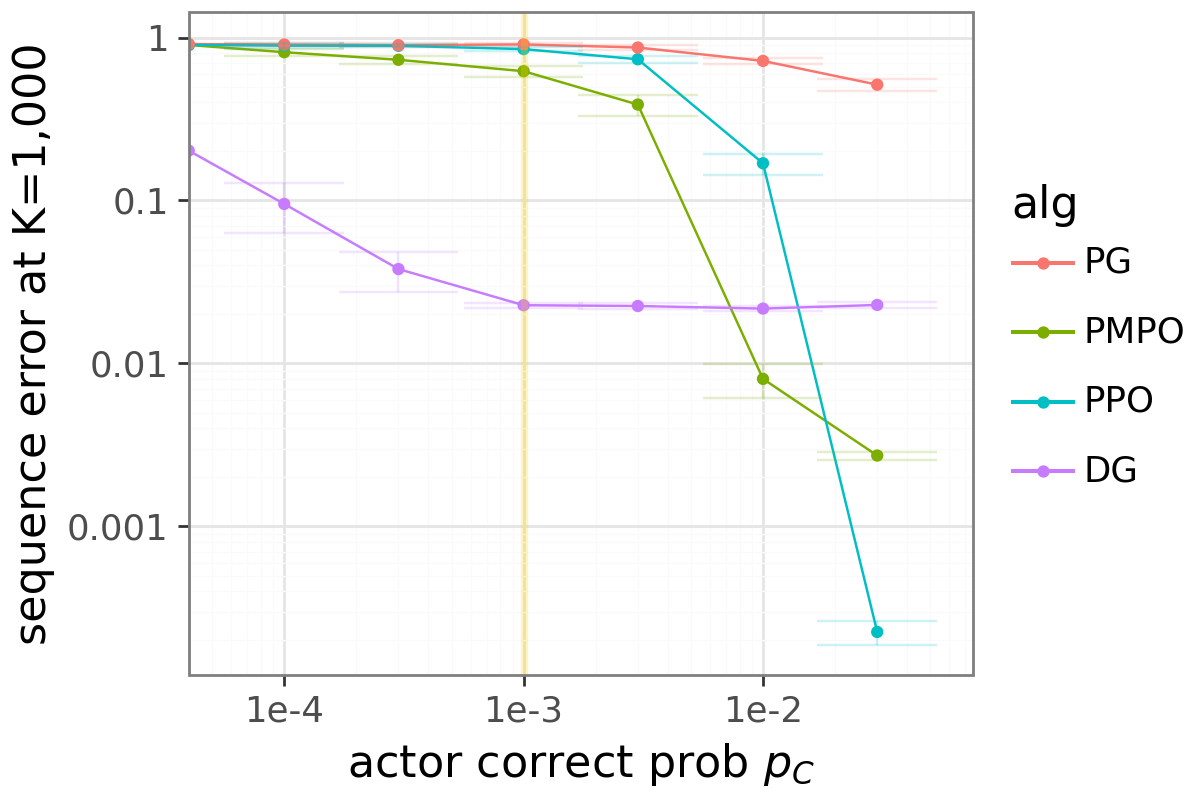}
    \vspace{-2mm}
    \caption{Sequence error at $K{=}1000$ vs.\ oracle rate $p_C$.}
    \label{fig:discovery_scaling}
\end{subfigure}
\vspace{-1mm}
\caption{Sensitivity to rare discovery under the hedonic trap ($\kappa{=}{-}1$). DG latches onto rare oracle trajectories; baselines require much higher oracle rates to make progress.}
\vspace{-1mm}
\label{fig:discovery}
\end{figure}

Figure~\ref{fig:discovery} shows that DG is the only method that consistently exploits these rare discoveries, and at $p_C{=}10^{-3}$ the only method that makes meaningful progress; PPO and PMPO need oracle rates near $10^{-2}$ or higher.

\FloatBarrier

\subsection{Combined Friction}
\label{sec:combo}

In practice, frictions occur together, so we ask whether DG's advantage composes.
We combine them at the operating points from the individual experiments: delay $D{=}30$, bug probability $p_E{=}3{\times}10^{-3}$, reward noise $p_R{=}0.01$, oracle rate $p_C{=}10^{-3}$, and the hedonic trap $\kappa{=}{-}1$.
The combination is tuned once at $H{=}5$; for each $H$ we then train a new model with these hyperparameters.
A length $H$ counts as solved once the mean training reward at that length first exceeds $0.5$, and $H^*$ is the largest $H$ such that every length up to $H$ is solved within a given number of gradient steps.
Curves average $H^*$ over 10 seeds, so the mean need not be an integer.

\begin{figure}[ht!]
\centering
\vspace{-1mm}
\begin{subfigure}[t]{0.48\columnwidth}
    \centering
    \includegraphics[width=\linewidth]{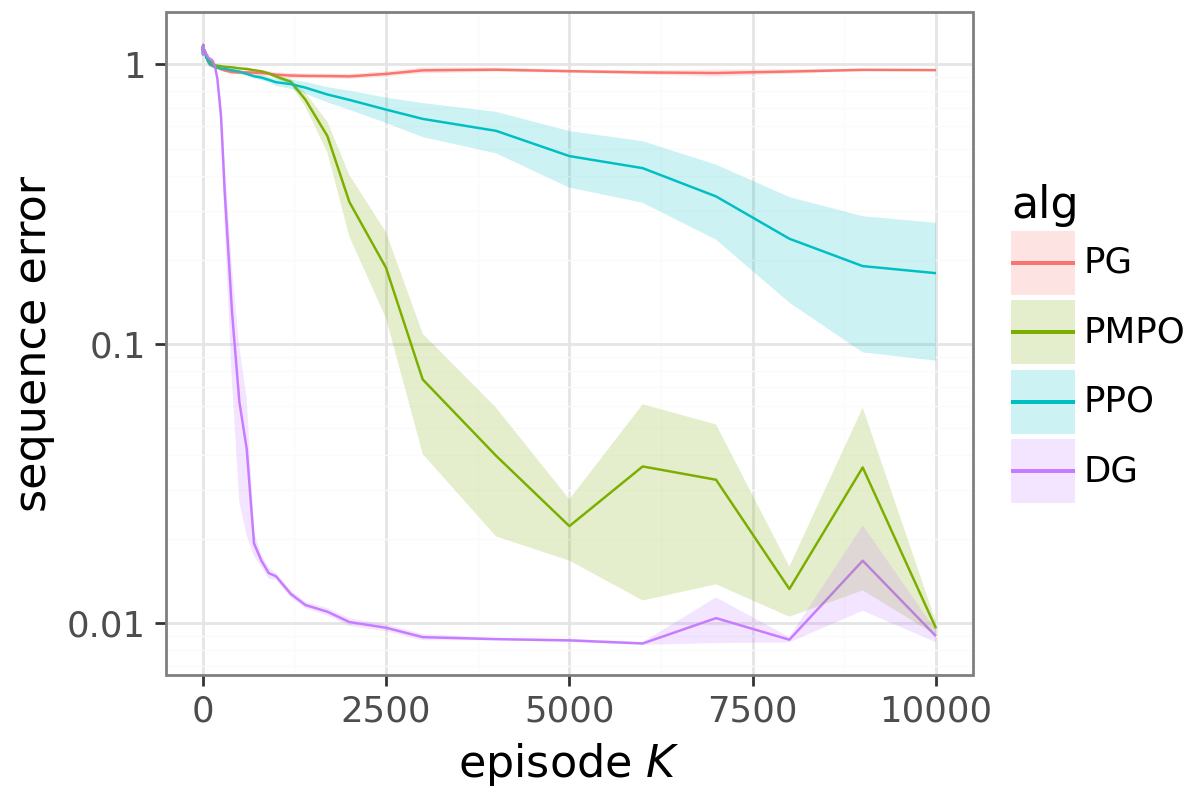}
    \vspace{-2mm}
    \caption{Learning curves at $H{=}5$.}
    \label{fig:combo_regret_5}
\end{subfigure}
\hfill
\begin{subfigure}[t]{0.48\columnwidth}
    \centering
    \includegraphics[width=\linewidth]{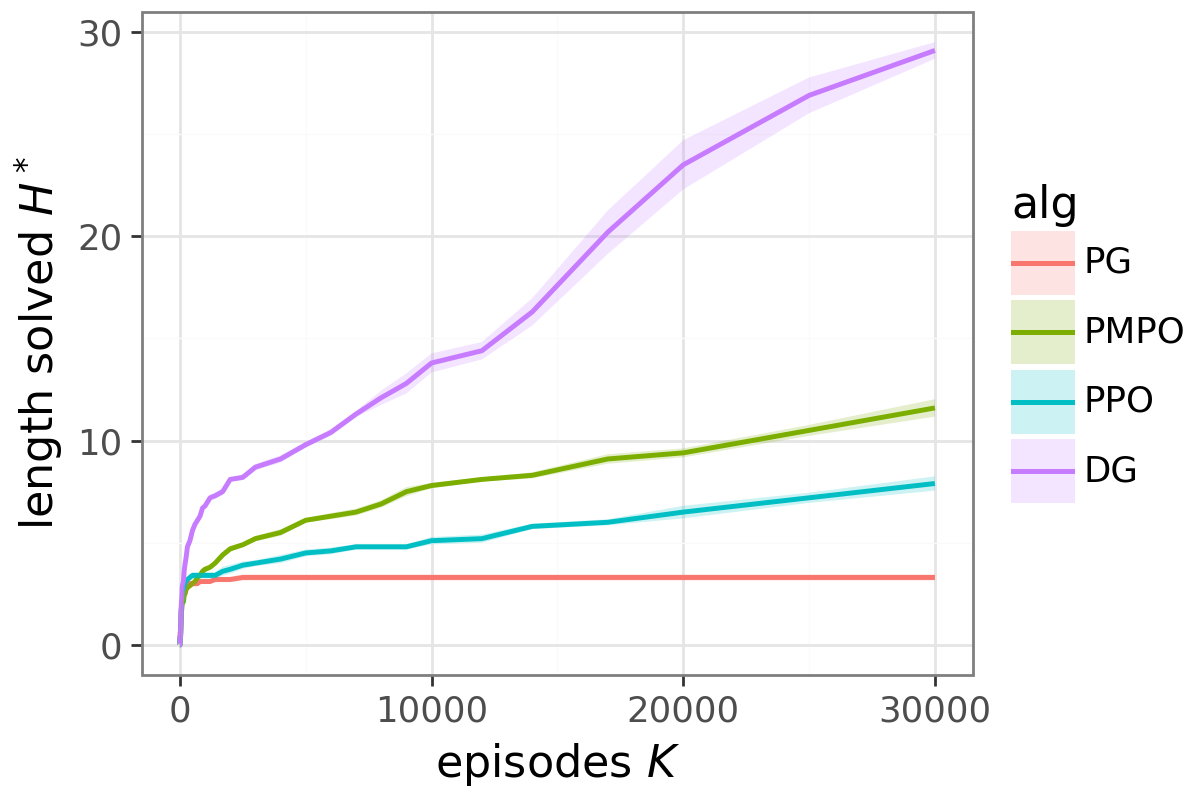}
    \vspace{-2mm}
    \caption{Largest sequence solved $H^*$ vs. gradient steps.}
    \label{fig:combo_scale}
\end{subfigure}
\vspace{-1mm}
\caption{Combined friction: staleness, actor bugs, reward corruption, and rare discovery at their single-friction operating points during training, scaling $H$.
DG solves longer sequences at each training budget; $H^*$ is measured from training reward with all frictions active.}
\vspace{-1mm}
\label{fig:combo}
\end{figure}

At $H{=}5$ (Figure~\ref{fig:combo}), DG reaches near-zero error within $2000$ gradient steps while PMPO requires $5000$ and PPO plateaus above $10\%$.
After $10\text{k}$ gradient steps, DG solves sequences up to $H^*{\approx}13$, compared to $H^*{\approx}8$ for PMPO, $H^*{\approx}5$ for PPO, and $H^*{\approx}3.5$ for PG.
Its advantage compounds with sequence complexity, consistent with the noise-suppression feedback from Section~\ref{sec:bandit}.

\FloatBarrier

\section{Related Work}
\label{sec:related}

\paragraph{Trust regions and off-policy correction.}
TRPO constrains updates through a KL trust region~\citep{schulman2015trust}, while PPO and GRPO use clipped probability ratios~\citep{schulman2017proximal,shao2024deepseekmath}.
Importance sampling corrects distribution mismatch when behavior probabilities are known~\citep{precup2001off}; V-trace, Retrace($\lambda$), and ACER truncate ratios to control variance~\citep{espeholt2018impala,munos2016safe,wang2017sample}.
These methods do not use the combination of learner-relative surprisal and advantage sign to distinguish rare successes from rare failures.
In the idealized tabular population setting, exact importance weighting recovers the on-policy expected gradient (Remark~\ref{rem:exact_is}).
DG can be preferable when behavior probabilities are unavailable, unreliable, or clipped, and when finite-batch learning benefits from sign-aware sample selection.

\paragraph{Distributed architectures.}
IMPALA, Ape-X, SEED, and Podracer reduce staleness through systems design, e.g.\ centralized inference, synchronous batching, or correction terms~\citep{espeholt2018impala,horgan2018distributed,espeholt2020seed,hessel2021podracer}.
DG is a complementary sample-weighting rule that can be inserted into these pipelines.

\paragraph{Filtered policy gradients, exploration, and offline RL.}
PMPO thresholds updates by advantage sign~\citep{abdolmaleki2024preference}; RWR and AWR weight by advantage~\citep{peters2007reinforcement,peng2019advantage}; and REINFORCE-leave-one-out improves baselines~\citep{kool2019buy}.
These methods are surprisal-blind and do not distinguish rare successes from rare failures under the learner's current policy.
Exploration methods generate novel data~\citep{bellemare2016unifying,pathak2017curiosity,ecoffet2021first}; DG addresses how to weight rare successes once they appear.
Offline RL methods constrain policies toward the data distribution~\citep{kumar2020conservative,kostrikov2021offline,chen2021decision}; DG instead filters data to improve update direction.

\section{Conclusion}
\label{sec:conclusion}

Distributed policy gradient learns from actors it does not control, and many of the actions that surprise the learner are blunders while a few are breakthroughs.
DG separates the two with a sigmoid gate on delight and needs no behavior probabilities.
Across MNIST, contaminated bandits, and token reversal, this learner-relative selection beats exact importance weighting and tuned PPO and PMPO, and its advantage grows with contamination and sequence length.
Off-policy correction asks how likely an actor was to take an action; delight asks what the action teaches the learner.

The experiments are controlled diagnostics that isolate individual mechanisms in small models; they include no end-to-end asynchronous actor--learner system and no literal inference-stack comparison.
DG is deliberately biased, and our theory characterizes that bias only in the analyzed bandit.
Its main failure mode is \emph{false delight}: high-surprisal positive-advantage samples caused by reward variance, verifier error, or reward hacking look like genuine discoveries.
\citet{osband2026doesgradientsparkjoy} makes this precise for a rare suboptimal action with reward gap $\Delta$ and action-specific reward noise $\sigma$: lucky draws produce positive delight, and DG collapses sharply near $\sigma \approx \Delta$ while PG degrades gradually.
Practical variants should clip advantages, cap surprisal, and add ratio corrections when reliable behavior probabilities exist.

\section*{Acknowledgements and Disclosures}

We thank Ben Van Roy and Satinder Singh for detailed feedback on earlier drafts,
and Niko Yasui for helpful feedback on the scope of the tabular analysis.
We are grateful to Raia Hadsell, Zoubin Ghahramani, Demis Hassabis, and Satinder Singh for building the research environment at Google DeepMind that made this work possible.
This work was funded by Google DeepMind, where the author is employed.
The author has no other competing interests to declare.

\bibliography{references}
\bibliographystyle{plainnat}



\newpage
\appendix

\section{MNIST Diagnostic}
\label{app:mnist_details}

This appendix supplements the MNIST experiments of Section~\ref{sec:mnist} with full experimental details, a learning rate robustness check, and a baseline sensitivity analysis.
The baseline analysis provides further evidence for the mechanism behind DG.

\subsection{Experimental Details}
\label{app:mnist_setup}

Each MNIST image $x$ is presented as a contextual bandit: the agent samples a label $a \in \{0,\dots,9\}$ from a softmax policy $\pi_\theta(a \mid x)$ and receives reward $r = \mathbb{I}\{a = y\}$; the true label $y$ is never observed.
The policy is a two-layer ReLU MLP with hidden width 100, trained with Adam (learning rate $10^{-3}$) over minibatches of $B{=}100$ images.
We validate every 500 gradient steps on the full 10{,}000-image test set with greedy decoding.
We use an oracle expected-reward baseline,
\[
b(x)=\mathbb{E}_{a\sim\pi_\theta(\cdot\mid x)}[r(a)]=\pi_\theta(y\mid x),
\]
which requires the true label and is therefore used only as a diagnostic control.
Because the learner never observes $y$ directly, this is oracle supervision used only for this diagnostic; it gives every method access to the best possible baseline, ensuring that performance differences reflect gradient weighting rather than baseline quality.

Staleness is modeled by a fixed delay: the actor uses the learner parameters from $D$ steps earlier, and the initial parameters for the first $D$ steps.
We sweep $D \in \{0, 1, 3, 10, 30, 100, 300, 1000\}$.
We compare three methods, all sharing architecture, optimizer, and baseline:
REINFORCE uses the stale-policy gradient without correction;
PG applies exact importance weighting $\pi_\theta(a \mid x) / \mu_{\theta'}(a \mid x)$ using the stored checkpoint that generated the action;
DG ($\eta{=}1$) weights updates by $\sigma(\text{delight}/\eta)$ using no importance weights and no knowledge of the actor's policy.
All results average over 30 seeds with $\pm 1$ standard error.
The learning rate was selected for best average performance across methods at $D{=}30$; we did not tune per-method.

\subsection{Learning Rate Robustness}
\label{app:mnist_lr}

\begin{figure}[ht!]
\centering
\begin{subfigure}[t]{0.48\columnwidth}
    \centering
    \includegraphics[width=\linewidth]{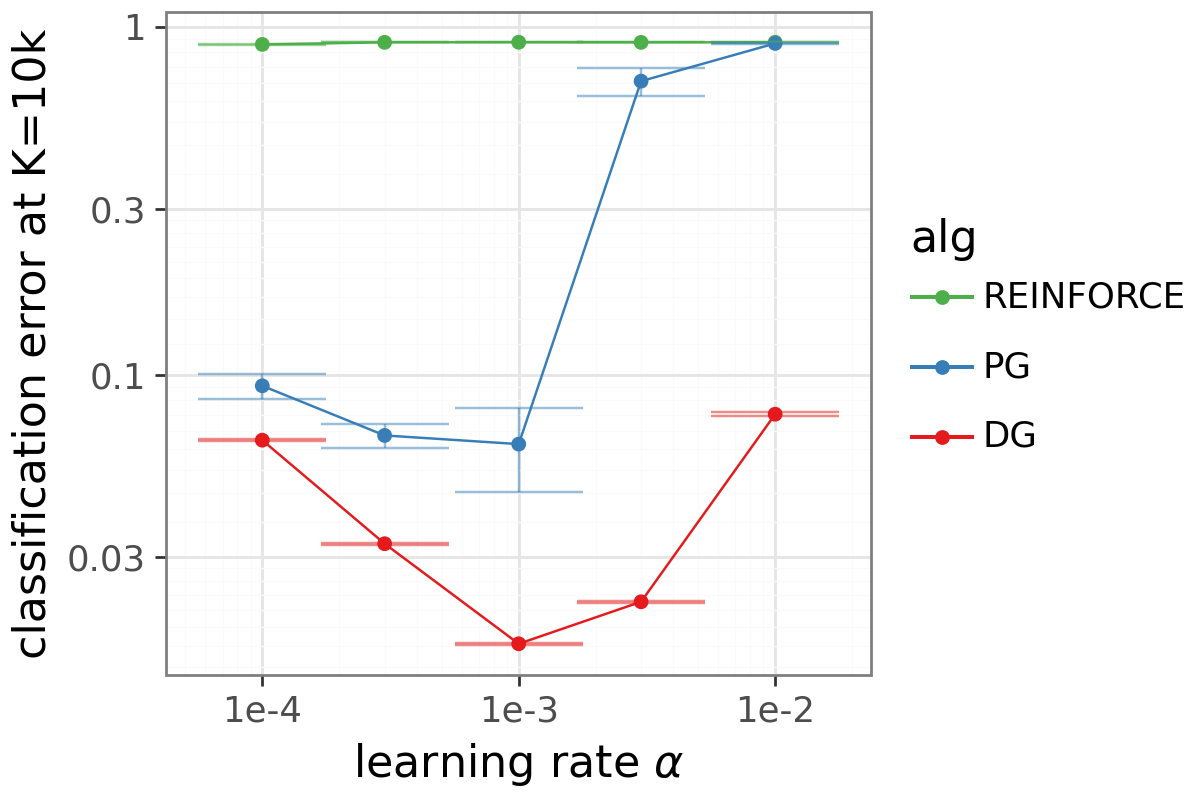}
    \caption{Training error vs.\ learning rate.}
    \label{fig:mnist_lr_train}
\end{subfigure}
\hfill
\begin{subfigure}[t]{0.48\columnwidth}
    \centering
    \includegraphics[width=\linewidth]{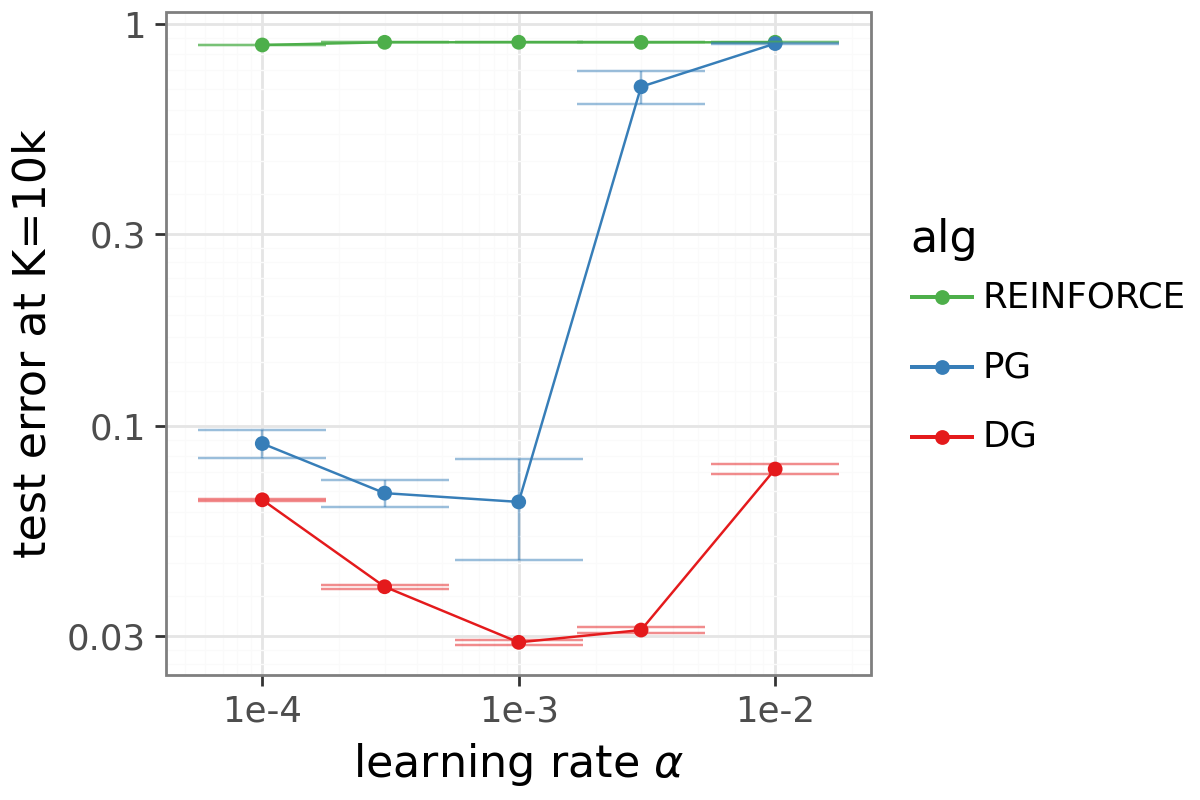}
    \caption{Test error vs.\ learning rate.}
    \label{fig:mnist_lr_test}
\end{subfigure}
\caption{Learning rate sweep on MNIST at $D{=}30$.
PG and DG perform best near $\mathrm{lr} = 10^{-3}$, with DG reaching lower error across the sweep.
Method ordering is the same on training and test error.}
\label{fig:mnist_lr}
\end{figure}

Figure~\ref{fig:mnist_lr} sweeps the learning rate across REINFORCE, PG, and DG at delay $D{=}30$.
PG and DG share an optimum near $\mathrm{lr} = 10^{-3}$, with DG ahead across the full range.
Training error~(a) and test error~(b) track almost identically, confirming that the DG advantage reflects better optimization, not overfitting.

\subsection{Baseline Sensitivity}
\label{app:mnist_baselines}

\begin{figure}[ht!]
\centering
\includegraphics[width=\columnwidth]{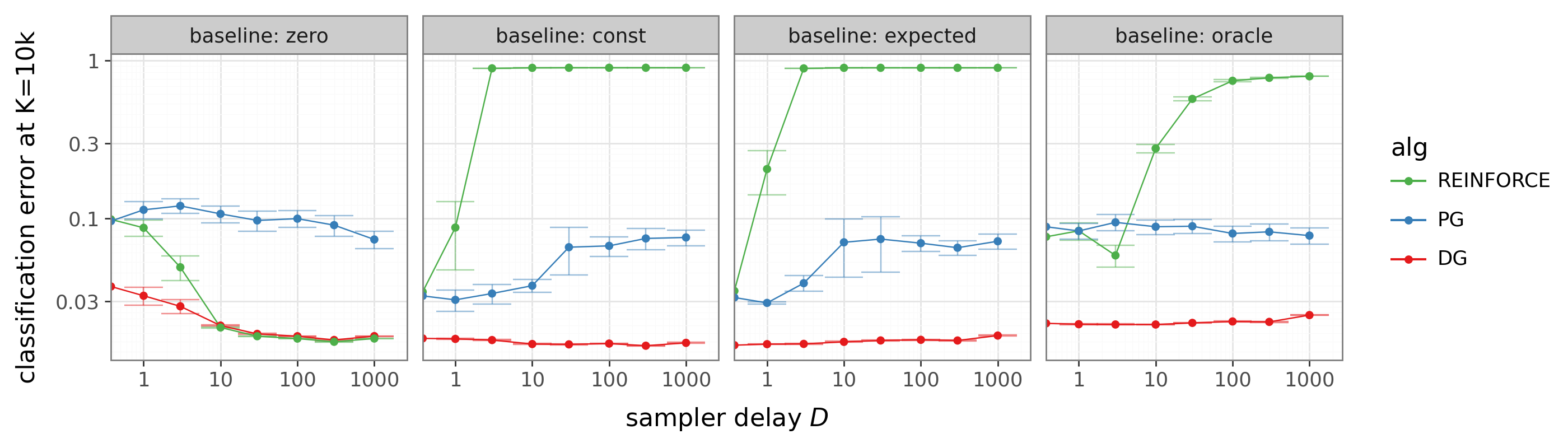}
\caption{Baseline sensitivity on MNIST.
Each panel sweeps sampler delay $D$ under a different baseline.
DG outperforms PG under all baselines.
Under the zero baseline, REINFORCE matches DG and PG is \emph{worse} than uncorrected REINFORCE.}
\label{fig:mnist_baselines}
\end{figure}

The main-text results use the oracle expected-reward baseline $b(x) = \sum_a \pi(a \mid x) \cdot r(a) = \pi_\theta(y\mid x)$.
To check whether DG's advantage depends on this choice, Figure~\ref{fig:mnist_baselines} repeats the staleness sweep under four baselines:
\emph{zero} ($b = 0$), \emph{constant} ($b = 0.5$), \emph{expected confidence} ($b = \sum_a \pi_\theta(a \mid x)^2$, which needs no label), and \emph{oracle} ($b = \mathbb{E}[R \mid x]=\pi_\theta(y \mid x)$, used in the main text).

DG outperforms PG across all four baselines.
The zero-baseline panel is the most revealing.
With $b = 0$, all advantages are non-negative ($U_t = R_t \ge 0$), so every nonzero update pushes toward the chosen action.
In this regime REINFORCE learns only from positive outcomes and its behavior closely tracks DG: both methods emphasize actions that yielded reward, regardless of whether the actor or learner generated them.
PG, by contrast, becomes \emph{worse} than uncorrected REINFORCE under the zero baseline.
Importance weighting can downweight off-policy successes precisely when the learner assigns them low probability, discarding the high-delight samples that DG and REINFORCE exploit.
This is consistent with DG's mechanism: when stale data contain a surprising success, learning from it works better than correcting it away.

\section{Proofs for Tabular Analysis}
\label{app:bandit_proofs}

This appendix proves the results of Section~\ref{sec:bandit}: the asymmetric tail gating lemma, perpendicular failure suppression, and the sign-blind tradeoff.
The analysis focuses on perpendicular second-moment suppression rather than population cosine collapse.
This distinction matters: in a single softmax bandit, a negative update on an incorrect action can still have a useful component in the true-gradient direction.
In the near-optimal regime, the absolute cosine between an incorrect-action score vector $\phi_\pi(a)$ and $\nabla_z J$ is $\Theta(1)$; equivalently, the negative update $-\phi_\pi(a)$ has positive $\Theta(1)$ cosine with the true-gradient direction.

All gradients are with respect to logits $z\in\mathbb{R}^K$.
Recall that $\phi_\pi(a):=e_a-\pi$ is the logit-space gradient of $\log\pi(a)$, and the true gradient is $\nabla_z J=\pi(y^*)\,\phi_\pi(y^*)$.
Let $u := \phi_\pi(y^*)/\|\phi_\pi(y^*)\|$ denote the unit true-gradient direction and $\Pi_\perp := I - uu^\top$ the orthogonal projection.
The advantage is $U(a)\in\{+1/2,-1/2\}$ and actions are sampled from the contaminated distribution $\mu=(1-\rho)\pi+\rho\,\nu$.

\begin{remark}[Stronger suppression at small $\eta$]
Write $\delta_{\max} := \max_{a\neq y^*}\pi(a)$.
When $\eta \le 1/2$, $\pi(a)^{1/\eta} \le \pi(a)^2$ for all disfavored $a$, so the overlap moment $M_{\nu,\eta}^\perp(\pi) = O(\delta_{\max}^2)$, faster than the $O(\delta_{\max})$ rate at $\eta=1$.
All experiments in this paper use $\eta = 1$.
\end{remark}

We work in the near-optimal regime $\pi(y^*)=1-\delta$ with $\delta\ll 1$.

\subsection{Proof of Lemma~\ref{lem:tail_gate}}

Let $p\in(0,1]$ and $c>0$.
For positive advantage,
\[
\sigma(c(-\log p)/\eta)
=
\frac{1}{1+p^{c/\eta}}
\ge
1-p^{c/\eta},
\]
where the final inequality uses $1/(1+x)\ge 1-x$ for $x\ge0$.
For negative advantage,
\[
\sigma(-c(-\log p)/\eta)
=
\frac{p^{c/\eta}}{1+p^{c/\eta}}
\le
p^{c/\eta}.
\]
\hfill$\square$

\subsection{Proof of Proposition~\ref{prop:perp_suppression}}

For a disfavored action $a\neq y^*$, the advantage is $U(a)=-1/2$.
Thus
\[
G_{\mathrm{RF}}(a)=-\tfrac12\phi_\pi(a),
\qquad
G_{\mathrm{DG}}(a)=-\tfrac12 w(a)\phi_\pi(a).
\]
Therefore
\[
\left\|\Pi_\perp G_{\mathrm{RF}}(a)\right\|^2
=
\tfrac14
\left\|\Pi_\perp\phi_\pi(a)\right\|^2 .
\]
For DG, Lemma~\ref{lem:tail_gate} with $c=1/2$ gives
$
w(a)
=
\sigma(-\ell(a)/(2\eta))
\le
\pi(a)^{1/(2\eta)}.
$
Squaring,
$
w(a)^2\le \pi(a)^{1/\eta}.
$
Hence
\[
\left\|\Pi_\perp G_{\mathrm{DG}}(a)\right\|^2
\le
\tfrac14
\pi(a)^{1/\eta}
\left\|\Pi_\perp\phi_\pi(a)\right\|^2 .
\]
Taking expectation over $a\sim\mu$ restricted to $a\neq y^*$ yields the claimed bounds.
Substituting $\mu=(1-\rho)\pi+\rho\nu$ gives the expanded form:
\[
V^{\mathrm{DG}}_{\perp,\mathrm{fail}}
\le
\tfrac14
(1-\rho)
\sum_{a\neq y^*}
\pi(a)^{1+1/\eta}
\left\|\Pi_\perp\phi_\pi(a)\right\|^2
+
\tfrac14
\rho
\underbrace{\sum_{a\neq y^*}
\nu(a)\pi(a)^{1/\eta}
\left\|\Pi_\perp\phi_\pi(a)\right\|^2}_{=\,M_{\nu,\eta}^\perp(\pi)} .
\]
The first term is the on-policy failure contribution; the second is the contaminated failure contribution, controlled by the overlap moment $M_{\nu,\eta}^\perp(\pi)$.
For $\eta=1$, using $\|\Pi_\perp\phi_\pi(a)\|^2 \le \|e_a-\pi\|^2 \le 2$:
\[
M_{\nu,1}^\perp(\pi)
\le
2\sum_{a\neq y^*}\nu(a)\pi(a)
\le
2\delta_{\max}
\to 0.
\]
By contrast, REINFORCE's contaminated perpendicular second moment is $\tfrac14\rho\sum_{a\neq y^*}\nu(a)\|\Pi_\perp\phi_\pi(a)\|^2$, which remains $\Omega(\rho)$ whenever $\nu$ places non-vanishing mass on actions with nonzero perpendicular component.
Writing $C^2 := \max_{a\neq y^*}\|\Pi_\perp\phi_\pi(a)\|^2 \le 2$, the contaminated DG term is at most $\tfrac14\rho\, C^2\delta_{\max}^{1/\eta}$, so the DG/REINFORCE contamination ratio vanishes as $O(\delta_{\max}^{1/\eta})$ near the optimum.
\hfill$\square$

\subsection{Proof of Proposition~\ref{prop:sign_blind_tradeoff}}

A learner-probability-only gate assigns the same weight $q(p)$ to two actions with the same learner probability $p$, regardless of whether their advantage is positive or negative.
If $q(p)\to0$ as $p\to0$, then a rare success and a rare failure are both suppressed: the gate cannot distinguish them.
If $\liminf_{p\to0} q(p)>0$, then rare failures are not suppressed by vanishing probability, and the contaminated perpendicular second moment remains $\Omega(\rho)$ under the nondegeneracy condition stated in Section~\ref{sec:bandit}.
DG is sign-dependent:
\[
w_+(p)=\sigma(c(-\log p)/\eta)
=
\frac{1}{1+p^{c/\eta}}
\to1,
\]
while
\[
w_-(p)=\sigma(-c(-\log p)/\eta)
=
\frac{p^{c/\eta}}{1+p^{c/\eta}}
\to0.
\]
Thus DG preserves rare successes while suppressing rare failures, which no learner-probability-only sign-blind gate can do.
\hfill$\square$

\begin{remark}[Exact importance weighting]
\label{rem:exact_is}
Exact importance weighting $f(a)=\pi(a)/\mu(a)$ reconstructs the on-policy expected gradient:
\[
\bar g_{\mathrm{IW}}
=
\mathbb{E}_{a\sim\mu}\!\left[\frac{\pi(a)}{\mu(a)}\,U(a)\,\phi_\pi(a)\right]
=
\mathbb{E}_{a\sim\pi}[U(a)\,\phi_\pi(a)]
=
\nabla_z J,
\]
because $\mathbb{E}_{a\sim\pi}[\phi_\pi(a)]=0$ and $U(a)=r(a)-1/2$.
Thus exact importance weighting removes contamination bias at the population level and recovers the on-policy expected gradient, but requires accurate behavior probabilities.
It does not eliminate finite-sample variance; in practice, large ratios can increase variance, while clipped ratios introduce bias.
The experiments in Sections~\ref{sec:mnist}--\ref{sec:results} test whether DG can be a better finite-batch update when behavior probabilities are unavailable, unreliable, or numerically fragile.
\end{remark}

The gate also induces a balanced-progress structure on positive-success terms.
For classification-style updates with zero baseline, a context with correct-action probability $p$ contributes expected DG direction
\[
p\,\sigma((-\log p)/\eta)\,\nabla\log p .
\]
Since $\nabla p = p\nabla\log p$, this is the gradient of the concave utility
\[
F_\eta(p)=\int_0^p \sigma((-\log s)/\eta)\,ds ,
\qquad
\nabla F_\eta(p)=\sigma((-\log p)/\eta)\nabla p .
\]
For $\eta=1$, $F_1(p)=\log(1+p)$.
Thus, on positive-success terms, DG can be viewed as replacing a linear success objective with a more balanced progress utility, reducing the dominance of already-solved contexts.
The negative-advantage terms are governed by the rare-failure suppression analysis above.

\section{Token Reversal Details}
\label{app:token_reversal_details}

This appendix supplements the token reversal experiments of Section~\ref{sec:results} with architecture and optimization details, per-friction hyperparameter sweeps, and a reward shaping sensitivity analysis.
The experimental code is released at \codeurl.
Each friction type is tuned at a single representative operating point on 10 validation seeds, then evaluated on 30 held-out seeds.

\subsection{Architecture and Optimization}
\label{app:architecture}

All methods use a causal decoder-only transformer implemented in Flax/JAX with 3 layers, 4 attention heads, embedding dimension 64, feed-forward dimension 128, pre-norm LayerNorm, ReLU activations, no bias terms, and learned position embeddings (${\approx}100\text{K}$ parameters).
The policy head is a linear layer mapping the final transformer output to logits over the vocabulary~$\mathcal{V}$.
We use Adam~\citep{kingma2014adam} with learning rate $10^{-4}$, selected via a sweep at the reference friction level for each experiment.
No learning rate schedule, weight decay, or gradient clipping is applied.

Each gradient step processes a batch of 100 episodes: 10 prompts with 10 sampled responses each.
The value baseline for each response is the mean reward across the 10 responses to the same prompt (grouped baseline), shared across all methods.
This is the group-mean baseline of GRPO~\citep{shao2024deepseekmath} without its standard-deviation normalization; unlike the leave-one-out baseline of \citet{kool2019buy}, it includes the scored response in the mean.

\paragraph{Method implementation.}
PG is implemented as the PPO clipped surrogate with $\varepsilon = 10^{9}$ (so no clipping ever triggers) and no KL penalty ($\beta_{\mathrm{KL}}{=}0$), reducing to the standard importance-weighted policy-gradient estimator with a grouped baseline.
In the MNIST diagnostic (Section~\ref{sec:mnist}), the uncorrected variant without importance weights is labeled REINFORCE.
PPO uses single-epoch updates (no data replay) with token-level clipped importance ratios and the same grouped baseline; the clip parameter~$\varepsilon$ is swept as described below.
PMPO uses $\gamma{=}10^{-9}$ for numerical stability in the dispreferred log-likelihood term and the same grouped baseline for preference classification.
All methods share the same architecture, optimizer, and baseline; they differ only in gradient weighting.

Exact behavior probabilities are unambiguous in the staleness experiments, where actors are stored learner checkpoints.
For intervention experiments, actor bugs and oracle trajectories are generated by an external mixture rather than the actor policy alone.
We therefore interpret PPO as a practical clipped-ratio baseline using stored actor probabilities, not as exact off-policy correction for intervention-generated samples.
The MNIST diagnostic includes exact behavior probabilities to isolate the exact-importance-weighting comparison in a setting where ratios are unambiguous.
A natural extension is to combine DG with clipped, V-trace-style, or exact mixture-ratio correction.

\paragraph{Compute.}
All experiments run on a single CPU machine.
Each token reversal experiment ($K{=}1000$ gradient steps, ${\approx}100\text{K}$-parameter transformer) completes in under two minutes.
The full experimental campaign (five friction types, four methods, six to eight hyperparameter settings each, 10 tuning seeds plus 30 evaluation seeds per configuration) requires approximately 200 CPU-hours total.

\subsection{Per-Friction Hyperparameter Sweeps}

For each friction type, we sweep DG temperature $\eta$ and PPO clip $\varepsilon$ over $\{0, 0.01, 0.03, 0.1, 0.3, 1, 3, 10, 30, 100, 1000\}$, PMPO~$\alpha \in \{0, 0.1, \ldots, 1\}$, and the KL coefficient $\beta \in \{0, 0.01, 0.1, 1\}$ for all three; $\eta{=}0$ denotes the hard gate $\mathbb{I}\{\chi_t > 0\}$.
The learning rate is $10^{-4}$ for all methods across all friction types.
We select the setting with lowest sequence error at $K{=}1000$ gradient steps ($K{=}10{,}000$ for combined friction) on 10 validation seeds, then evaluate on 30 held-out seeds; DG's main-text results use $\eta{=}1$.
The plots below show sequence error averaged over 10 seeds (lower is better); the legend $\beta$ is the KL coefficient, swept over $\{0, 0.01, 0.1, 1\}$.

\begin{figure}[ht!]
\centering
\begin{subfigure}[t]{0.32\columnwidth}
    \includegraphics[width=\linewidth]{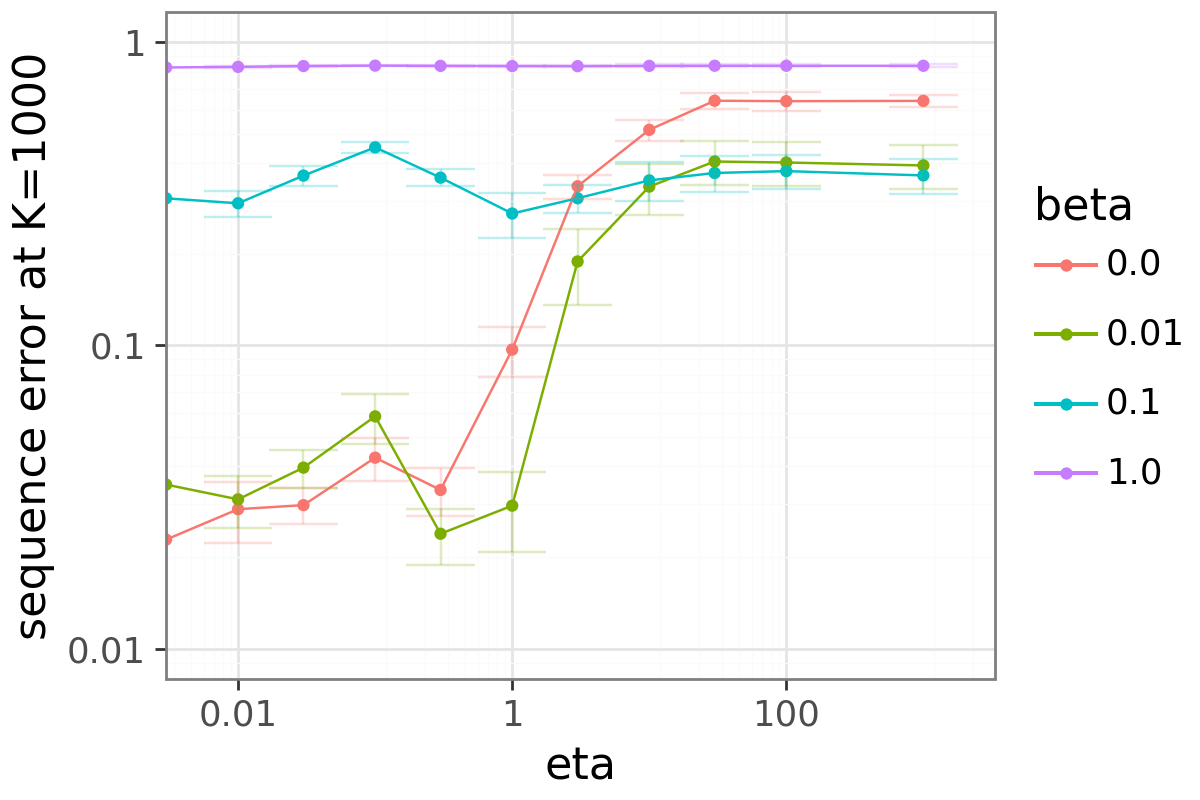}
    \vspace{-2mm}
    \caption{DG: $\eta$ sweep.}
\end{subfigure}
\hfill
\begin{subfigure}[t]{0.32\columnwidth}
    \includegraphics[width=\linewidth]{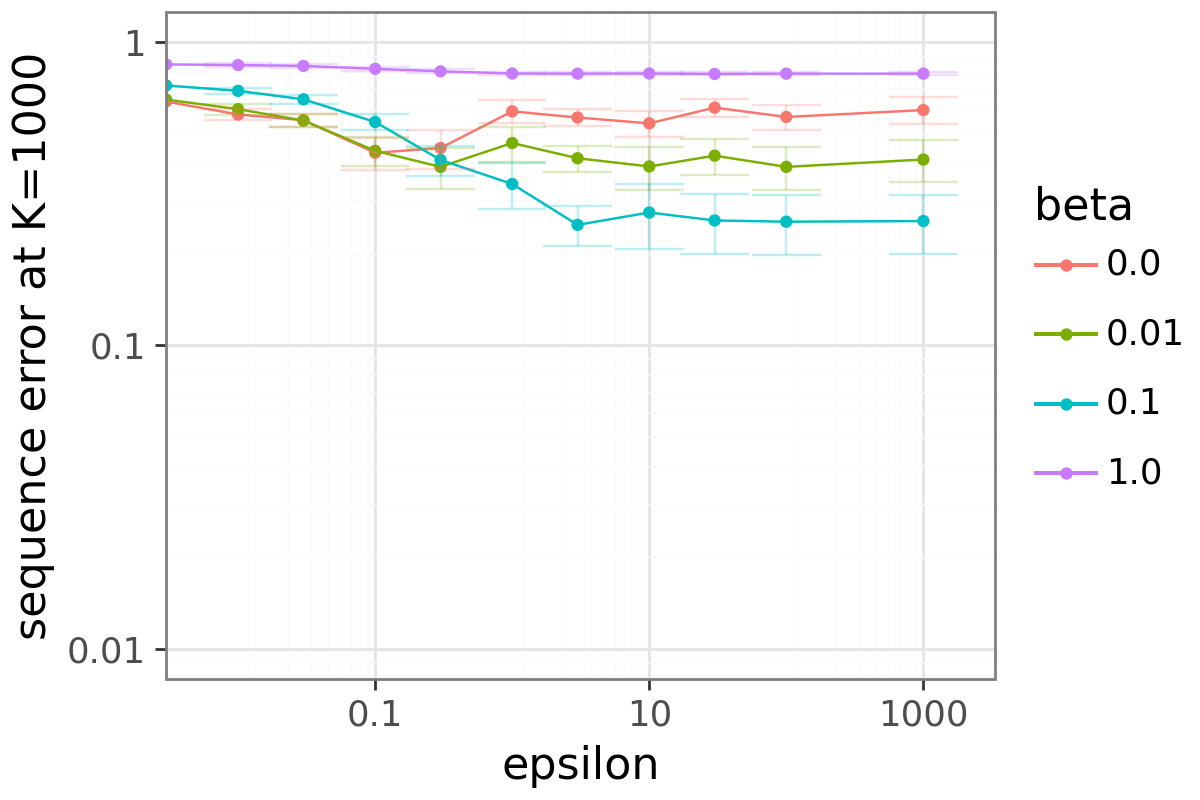}
    \vspace{-2mm}
    \caption{PPO: $\varepsilon$ sweep.}
\end{subfigure}
\hfill
\begin{subfigure}[t]{0.32\columnwidth}
    \includegraphics[width=\linewidth]{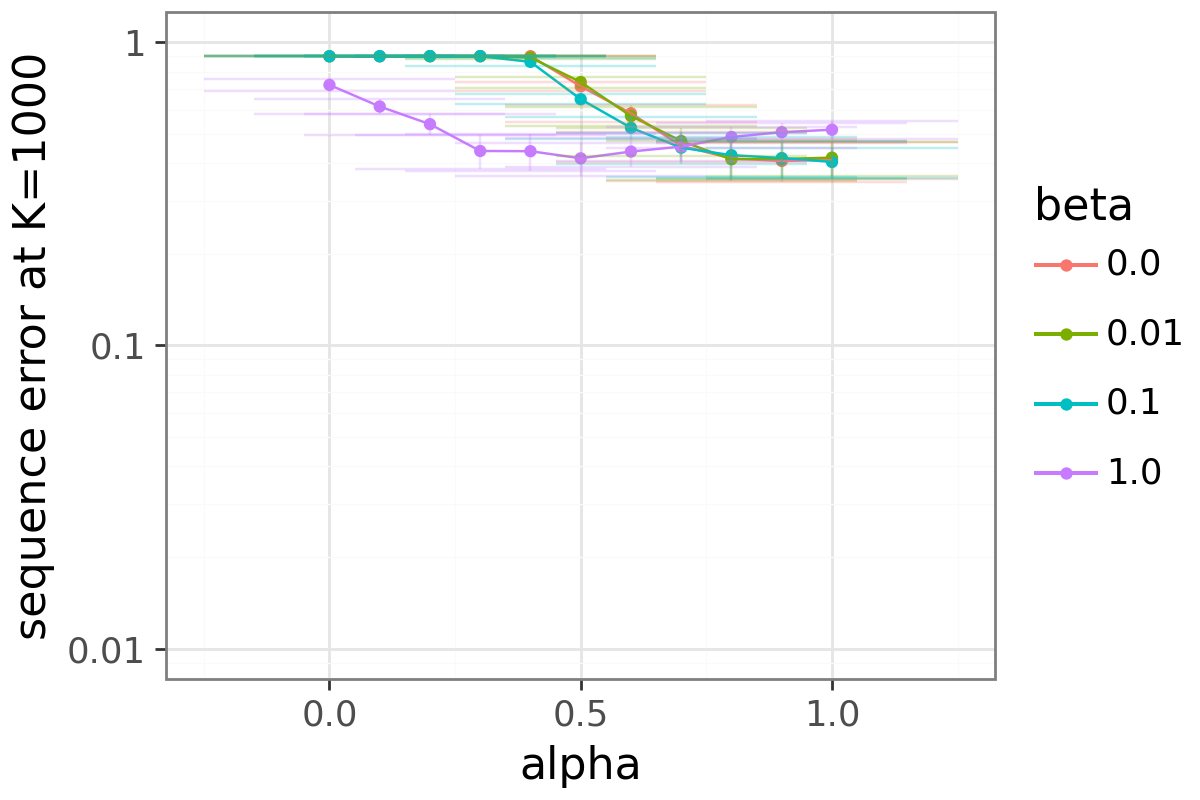}
    \vspace{-2mm}
    \caption{PMPO: $\alpha$ sweep.}
\end{subfigure}
\caption{Hyperparameter sensitivity under staleness ($D{=}30$), 10 seeds.}
\label{fig:delay_tune}
\end{figure}

\paragraph{Staleness (Section~\ref{sec:staleness}).}
Figure~\ref{fig:delay_tune} shows the sweep at $D{=}30$.
DG achieves low sequence error across a broad range of $\eta$; PPO and PMPO are less sensitive to their hyperparameters but never match DG.

\begin{figure}[ht!]
\centering
\begin{subfigure}[t]{0.32\columnwidth}
    \includegraphics[width=\linewidth]{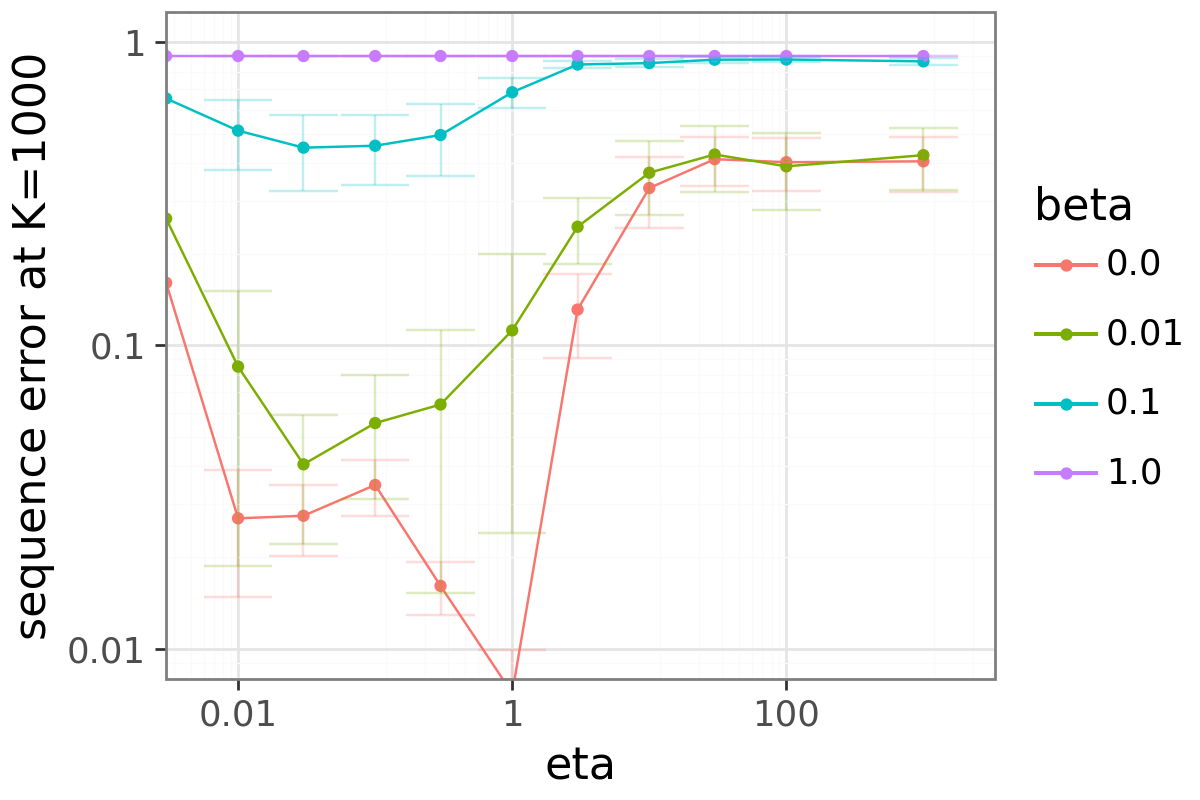}
    \vspace{-2mm}
    \caption{DG: $\eta$ sweep.}
\end{subfigure}
\hfill
\begin{subfigure}[t]{0.32\columnwidth}
    \includegraphics[width=\linewidth]{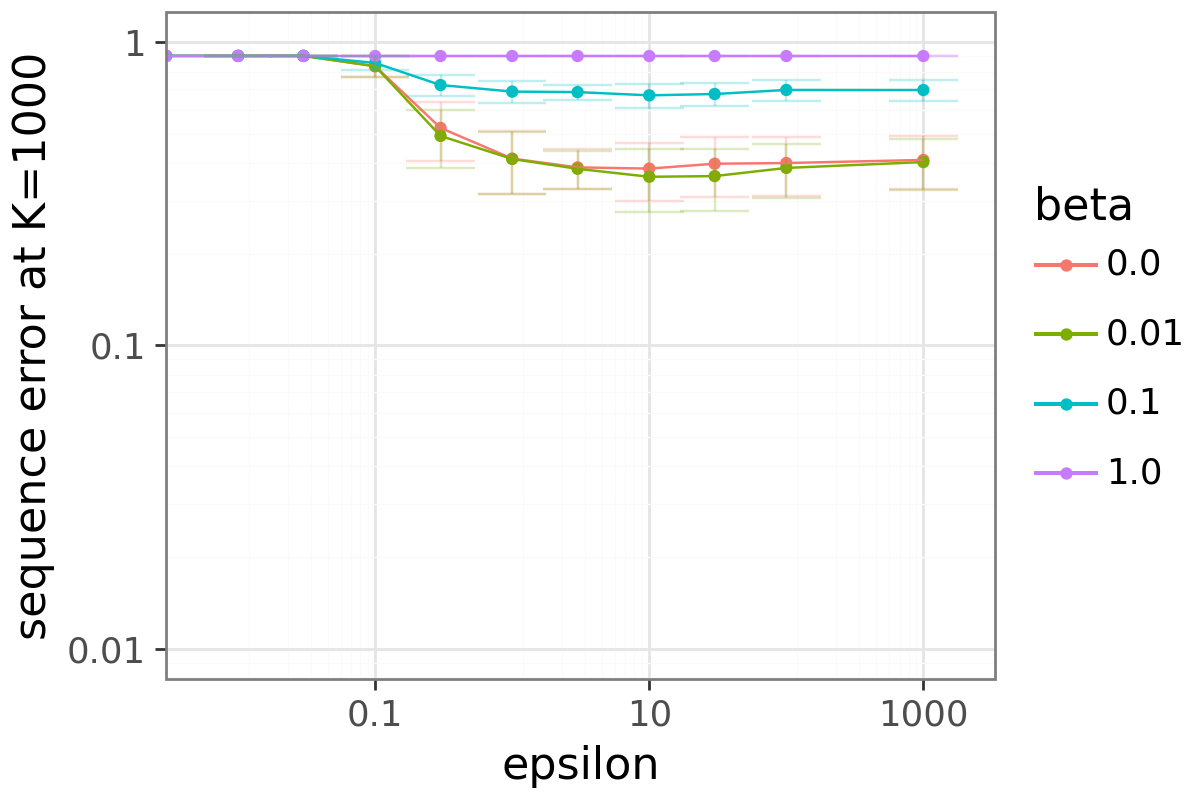}
    \vspace{-2mm}
    \caption{PPO: $\varepsilon$ sweep.}
\end{subfigure}
\hfill
\begin{subfigure}[t]{0.32\columnwidth}
    \includegraphics[width=\linewidth]{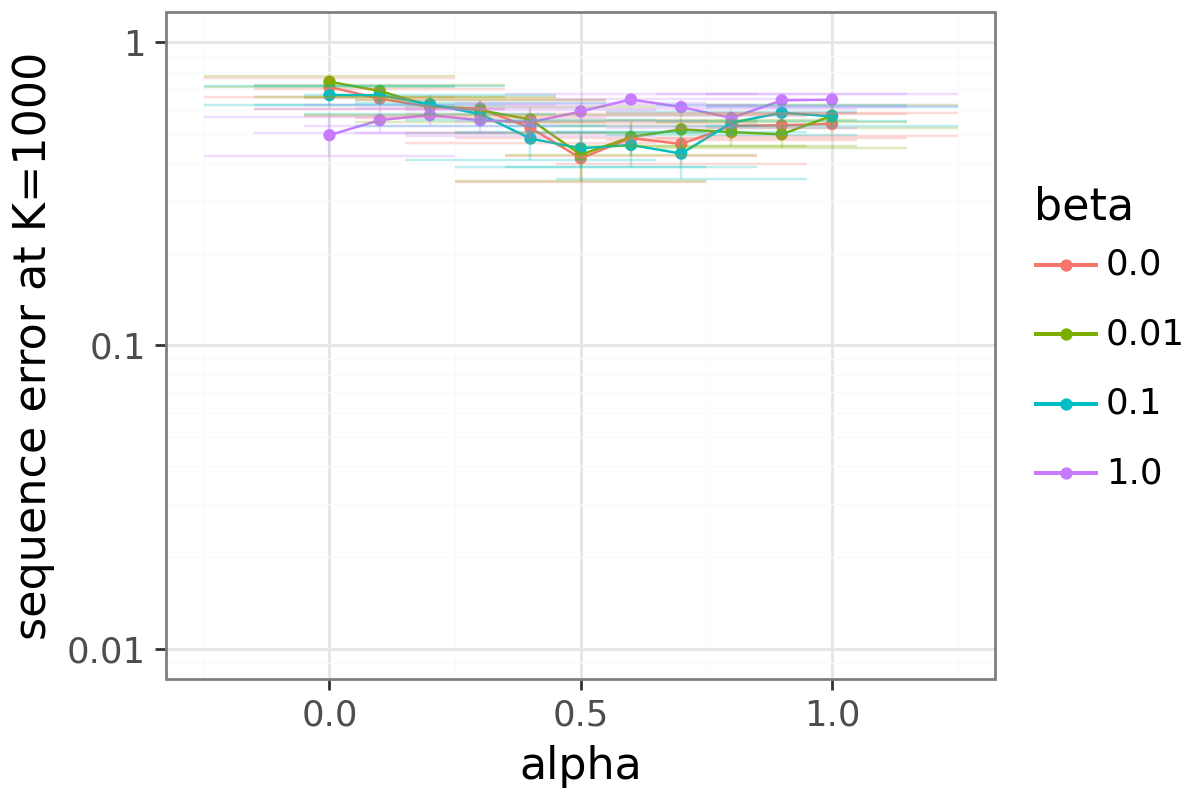}
    \vspace{-2mm}
    \caption{PMPO: $\alpha$ sweep.}
\end{subfigure}
\caption{Hyperparameter sensitivity under probability mismatch ($\sigma{=}0.3$), 10 seeds.}
\label{fig:mismatch_tune}
\end{figure}

\paragraph{Probability mismatch (Section~\ref{sec:mismatch}).}
Figure~\ref{fig:mismatch_tune} shows the sweep at $\sigma{=}0.3$.
DG is best at its default $\eta{=}1$ with no KL penalty, and strong KL regularization ($\beta \ge 0.1$) prevents learning; no PPO or PMPO configuration gets below $30\%$ error.

\begin{figure}[ht!]
\centering
\begin{subfigure}{0.32\columnwidth}
    \includegraphics[width=\linewidth]{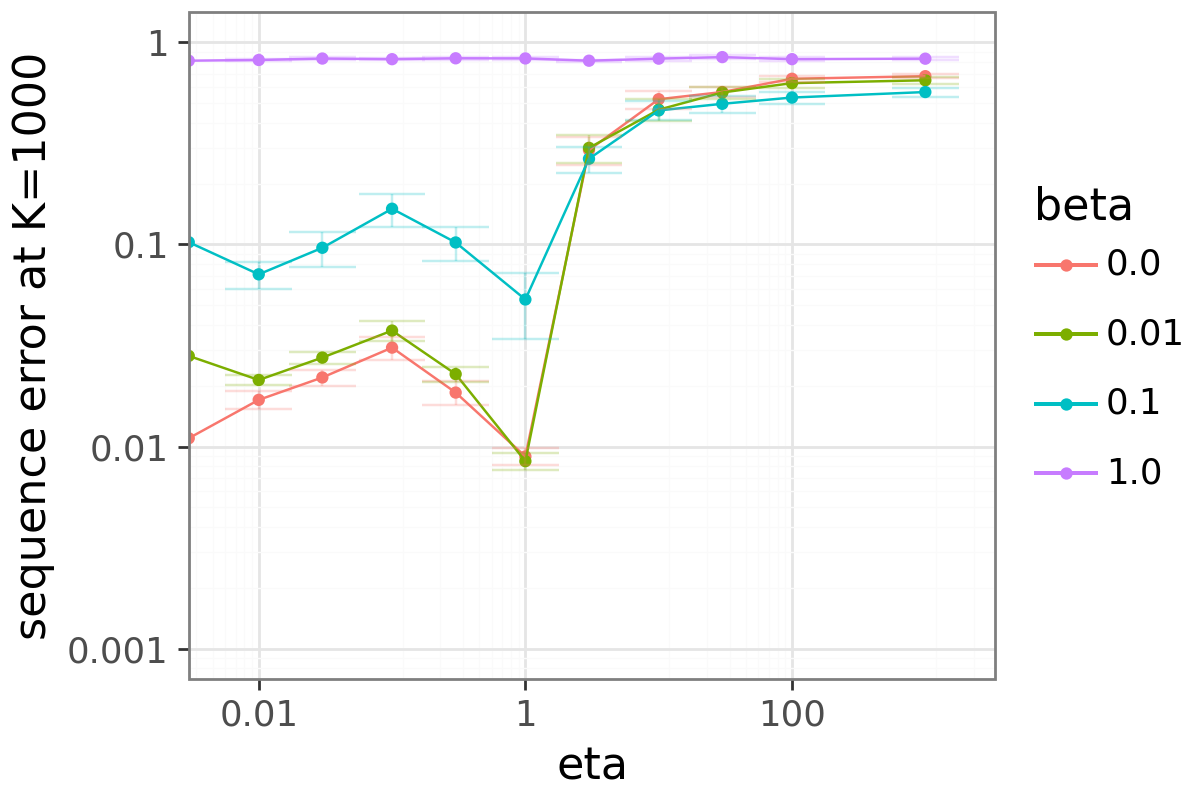}
    \vspace{-2mm}
    \caption{DG: $\eta$ sweep.}
\end{subfigure}
\hfill
\begin{subfigure}[t]{0.32\columnwidth}
    \includegraphics[width=\linewidth]{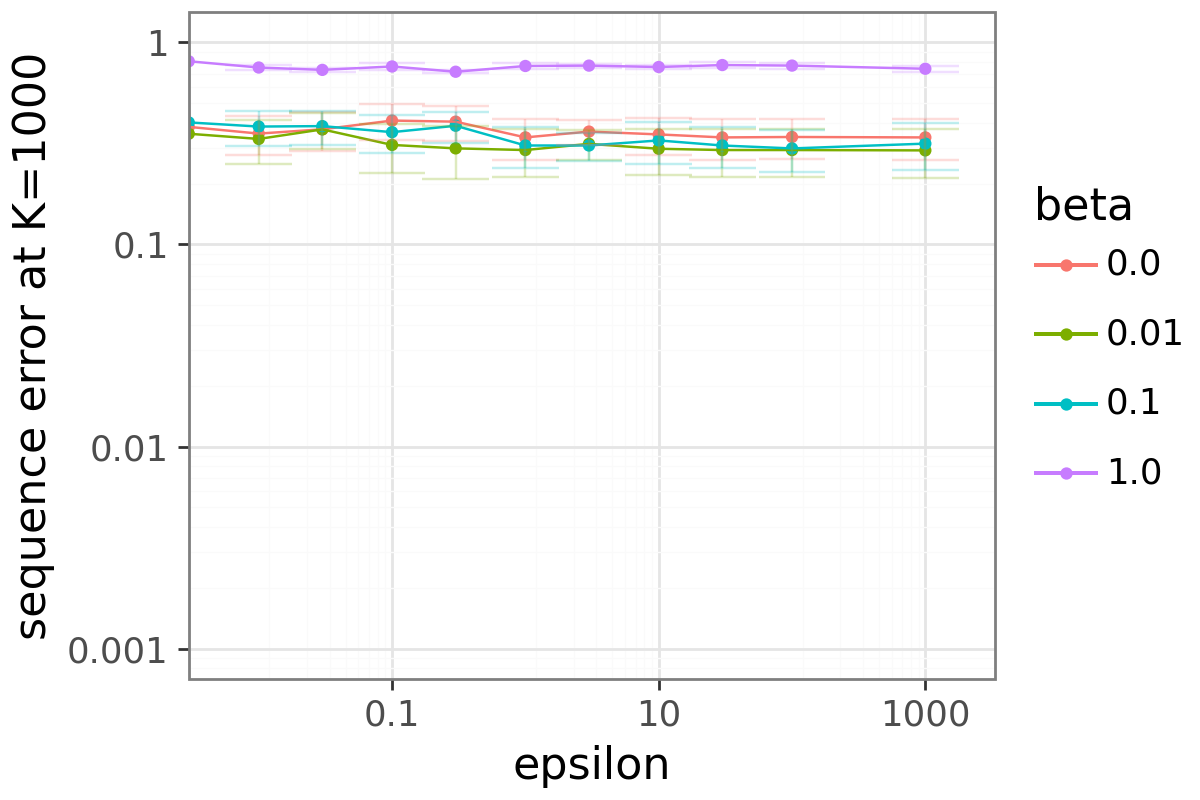}
    \vspace{-2mm}
    \caption{PPO: $\varepsilon$ sweep.}
\end{subfigure}
\hfill
\begin{subfigure}[t]{0.32\columnwidth}
    \includegraphics[width=\linewidth]{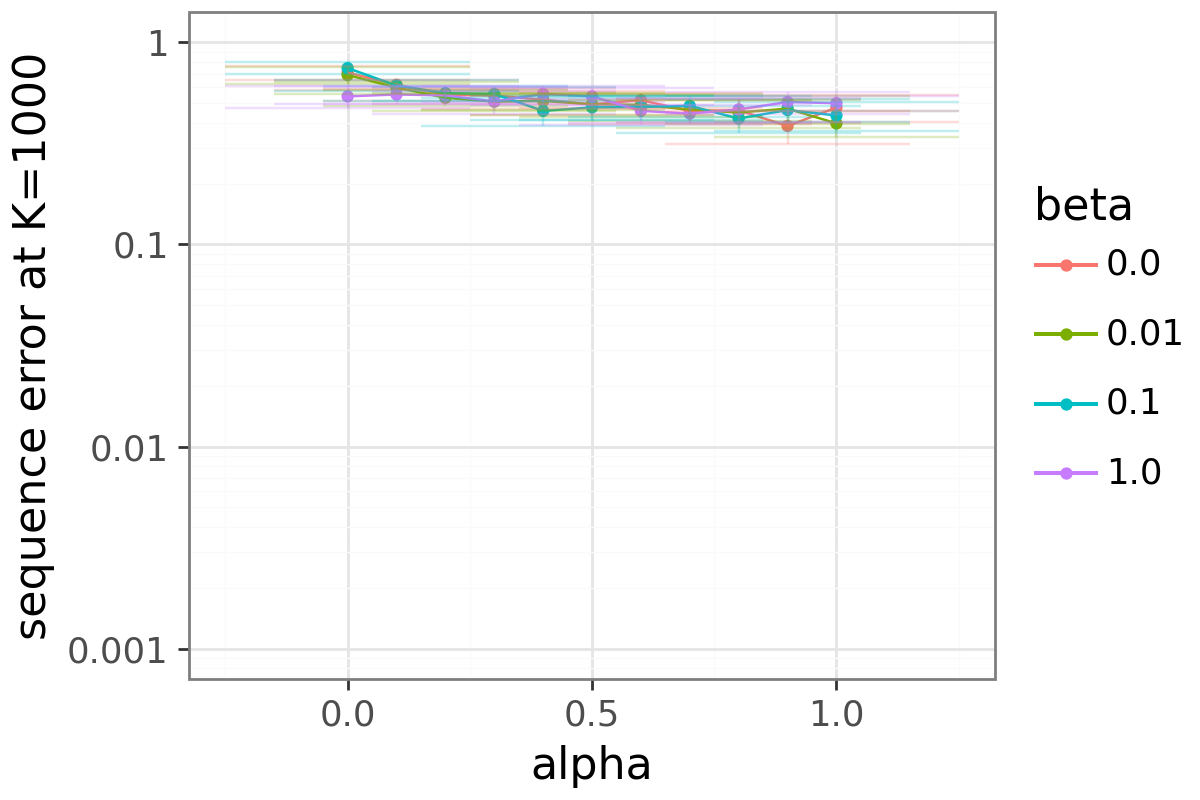}
    \vspace{-2mm}
    \caption{PMPO: $\alpha$ sweep.}
\end{subfigure}
\caption{Hyperparameter sensitivity under actor bugs ($p_E{=}3{\times}10^{-3}$), 10 seeds.}
\label{fig:bug_tune}
\end{figure}

\paragraph{Actor Bugs (Section~\ref{sec:bugs}).}
Figure~\ref{fig:bug_tune} shows the sweep at $p_E{=}3{\times}10^{-3}$.

\begin{figure}[ht!]
\centering
\begin{subfigure}[t]{0.32\columnwidth}
    \includegraphics[width=\linewidth]{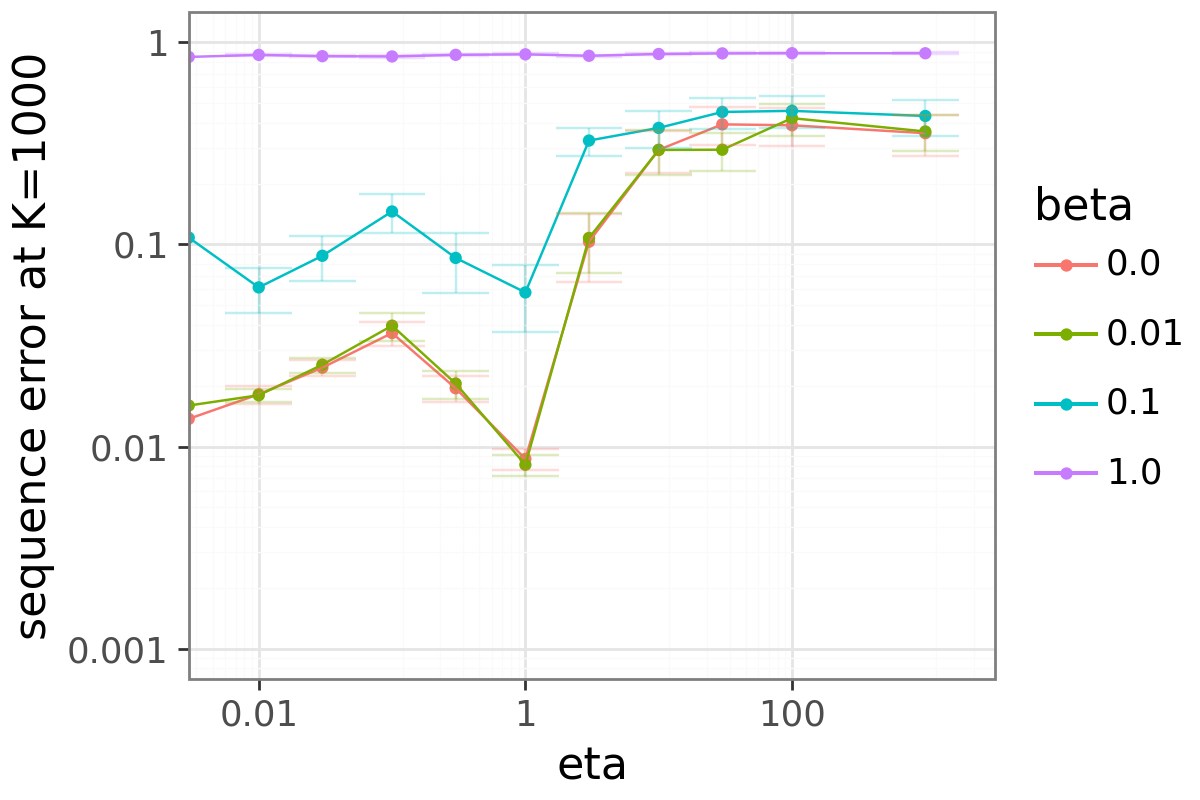}
    \vspace{-2mm}
    \caption{DG: $\eta$ sweep.}
\end{subfigure}
\hfill
\begin{subfigure}[t]{0.32\columnwidth}
    \includegraphics[width=\linewidth]{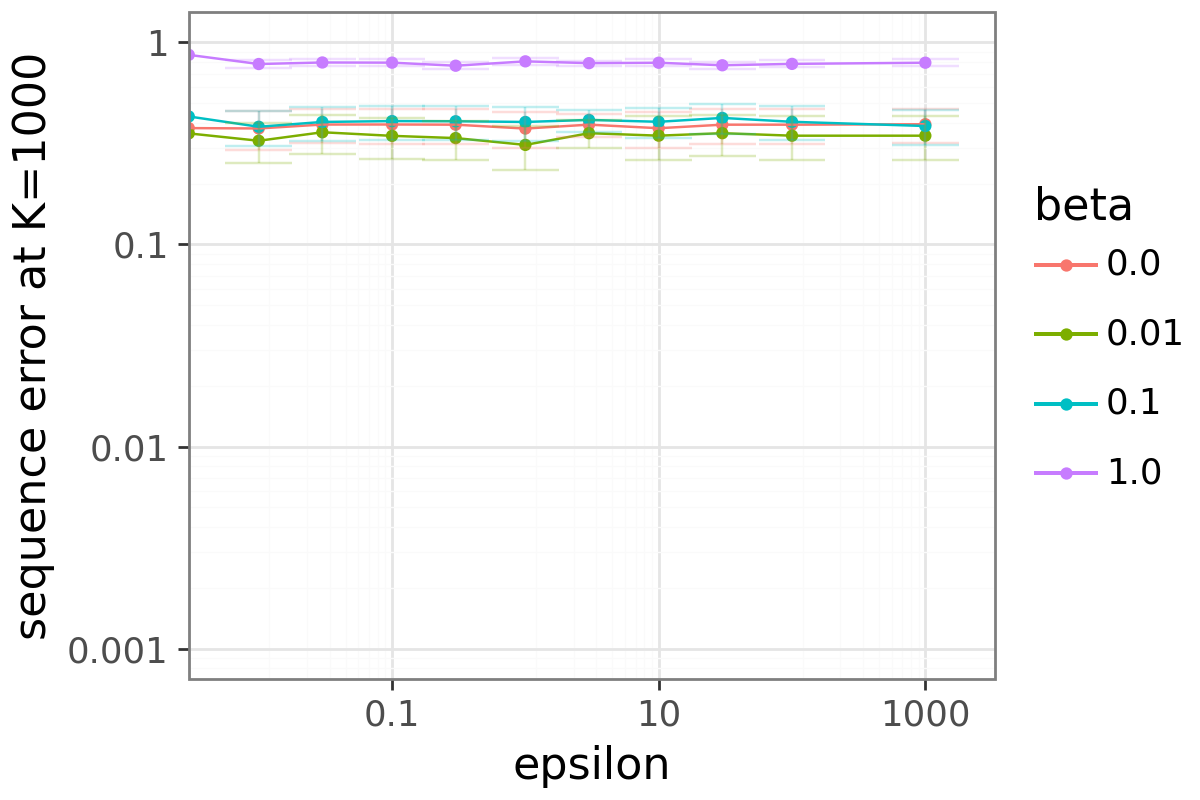}
    \vspace{-2mm}
    \caption{PPO: $\varepsilon$ sweep.}
\end{subfigure}
\hfill
\begin{subfigure}[t]{0.32\columnwidth}
    \includegraphics[width=\linewidth]{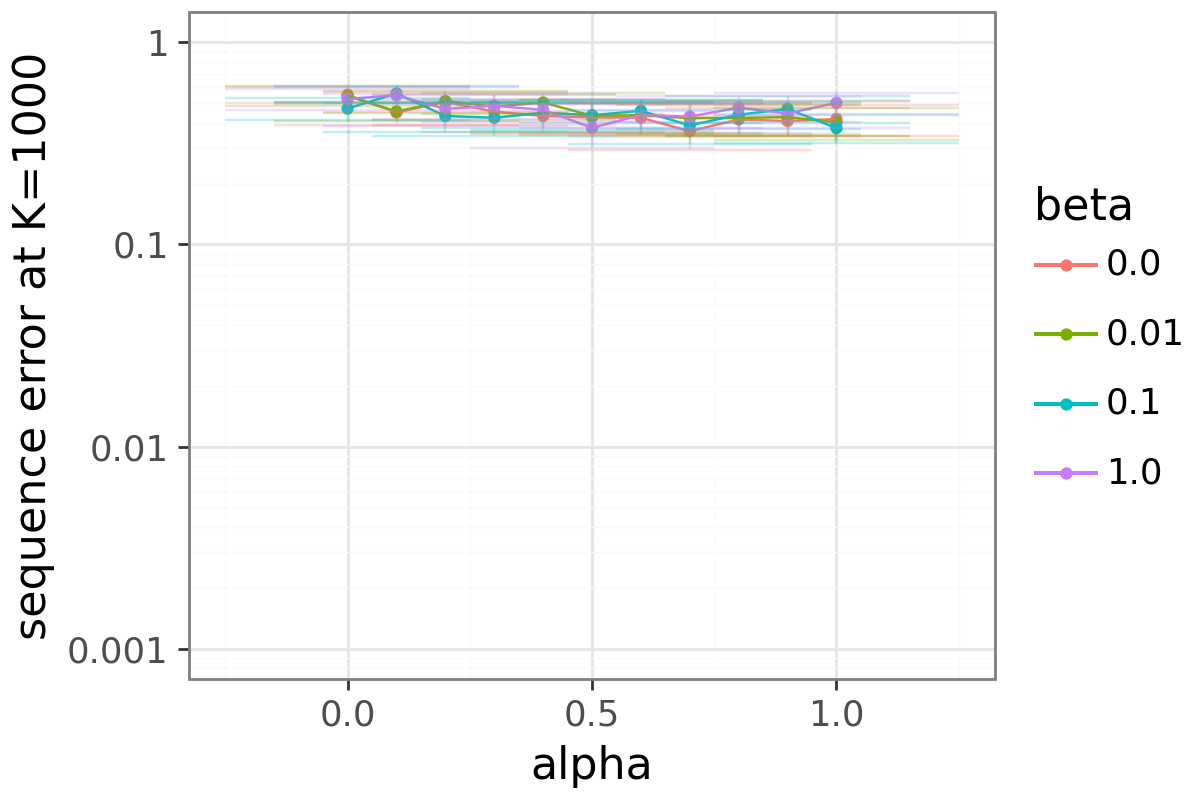}
    \vspace{-2mm}
    \caption{PMPO: $\alpha$ sweep.}
\end{subfigure}
\caption{Hyperparameter sensitivity under reward corruption ($p_R{=}0.01$), 10 seeds.}
\label{fig:rnoise_tune}
\end{figure}

\paragraph{Reward Corruption (Section~\ref{sec:reward}).}
Figure~\ref{fig:rnoise_tune} shows the sweep at $p_R{=}0.01$.

\begin{figure}[ht!]
\centering
\vspace{-2mm}
\begin{subfigure}[t]{0.32\columnwidth}
    \includegraphics[width=\linewidth]{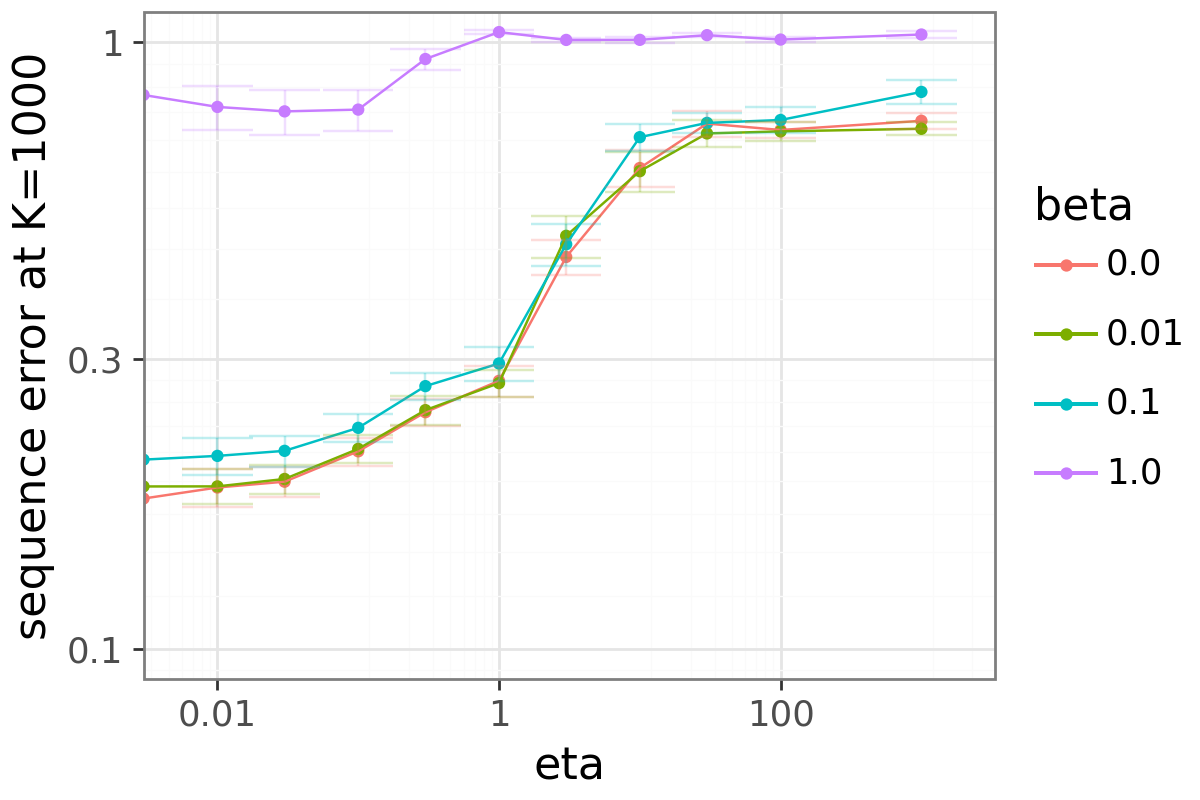}
    \vspace{-2mm}
    \caption{DG: $\eta$ sweep.}
\end{subfigure}
\hfill
\begin{subfigure}[t]{0.32\columnwidth}
    \includegraphics[width=\linewidth]{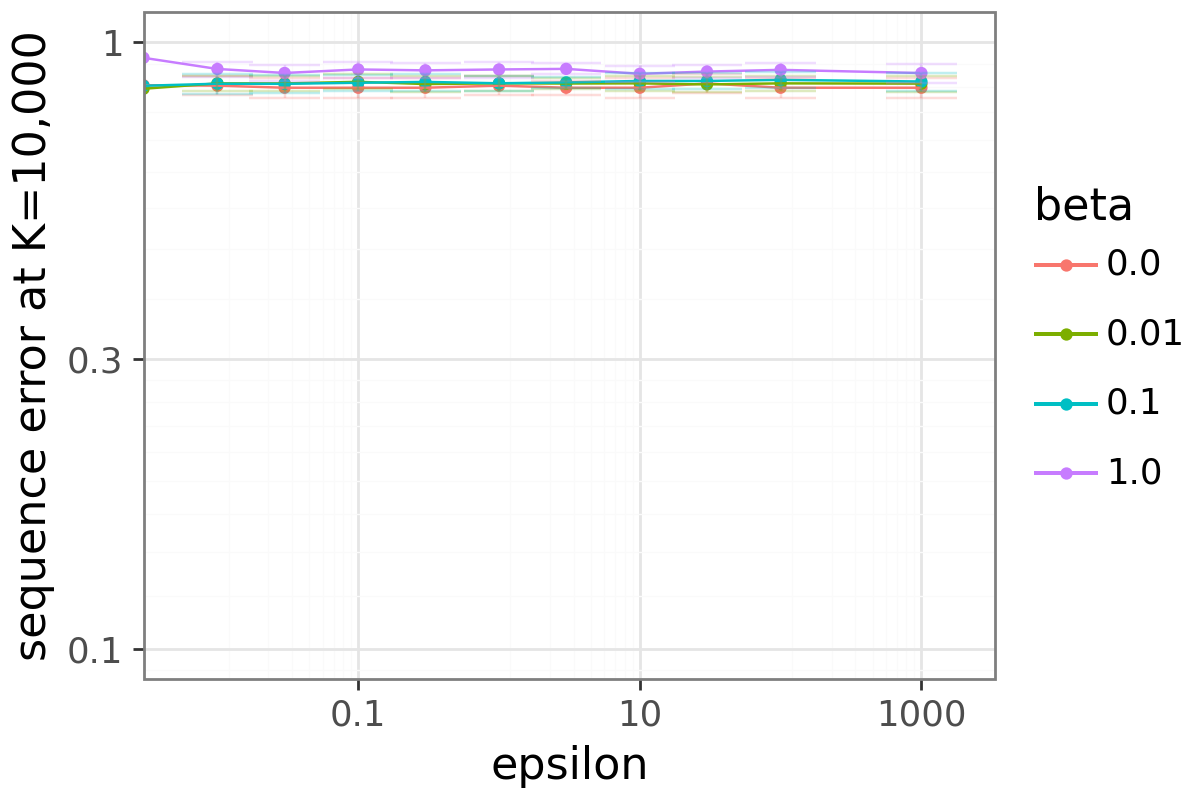}
    \vspace{-2mm}
    \caption{PPO: $\varepsilon$ sweep.}
\end{subfigure}
\hfill
\begin{subfigure}[t]{0.32\columnwidth}
    \includegraphics[width=\linewidth]{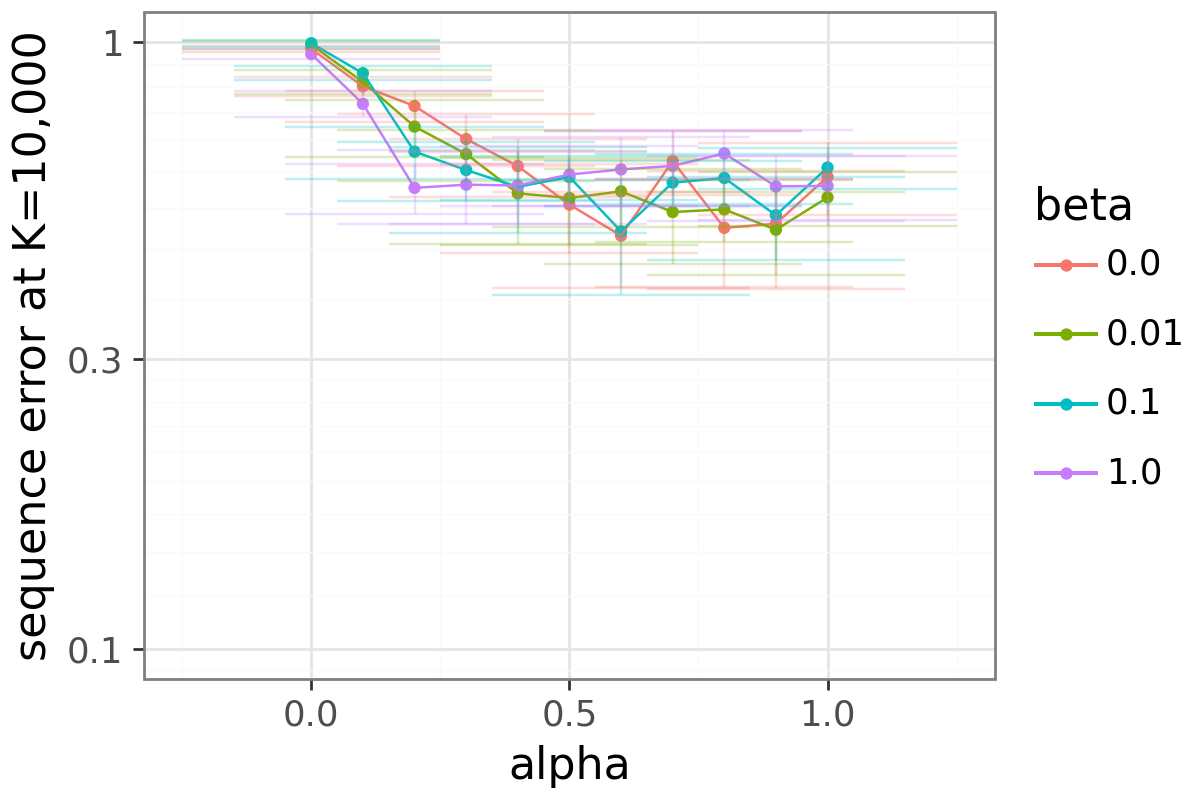}
    \vspace{-2mm}
    \caption{PMPO: $\alpha$ sweep.}
\end{subfigure}
\vspace{-2mm}
\caption{Hyperparameter sensitivity under rare discovery ($p_C{=}10^{-3}$), 10 seeds.}
\label{fig:explore_tune}
\vspace{-2mm}
\end{figure}

\paragraph{Rare Discovery (Section~\ref{sec:discovery}).}
Figure~\ref{fig:explore_tune} shows the sweep at $p_C{=}10^{-3}$.
This is the hardest setting: only DG finds configurations that reach low sequence error.
The best DG temperature is again near $\eta{=}1$, confirming that the default works across friction types.
No PPO or PMPO configuration makes meaningful progress at this oracle rate.

\begin{figure}[ht!]
\centering
\vspace{-2mm}
\begin{subfigure}[t]{0.32\columnwidth}
    \includegraphics[width=\linewidth]{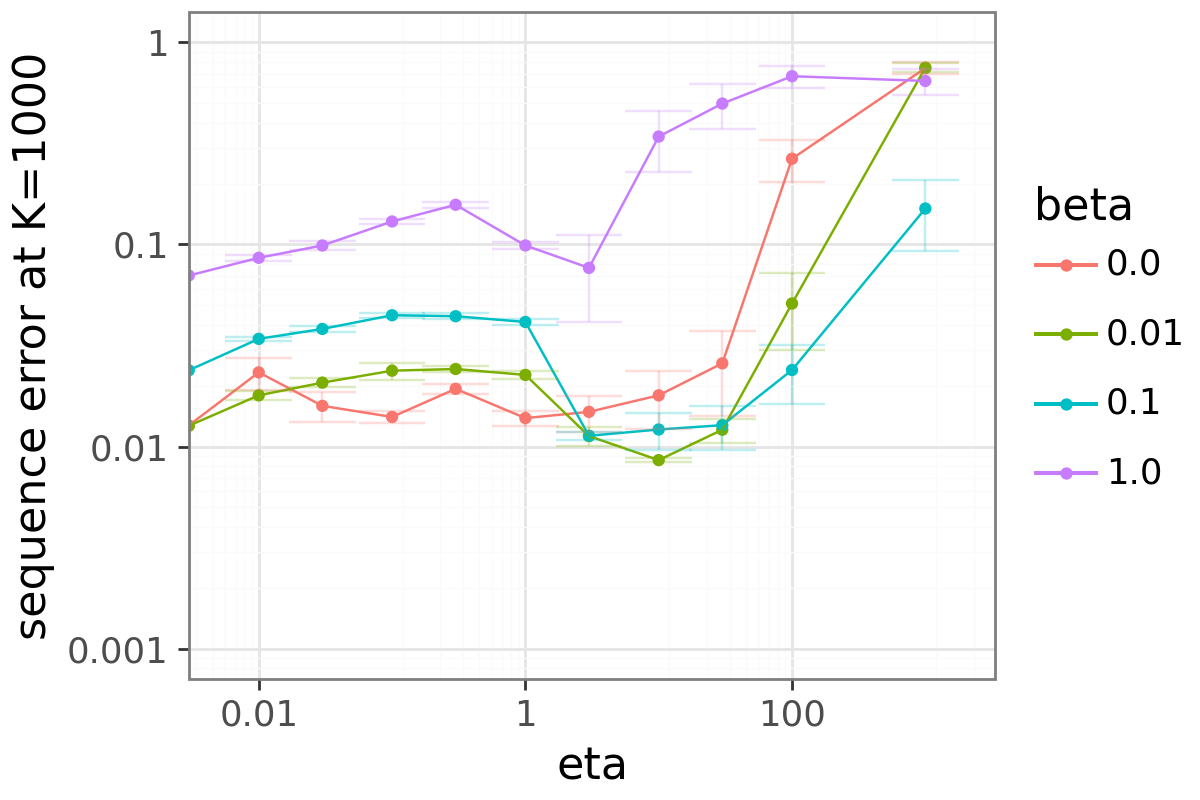}
    \vspace{-2mm}
    \caption{DG: $\eta$ sweep.}
\end{subfigure}
\hfill
\begin{subfigure}[t]{0.32\columnwidth}
    \includegraphics[width=\linewidth]{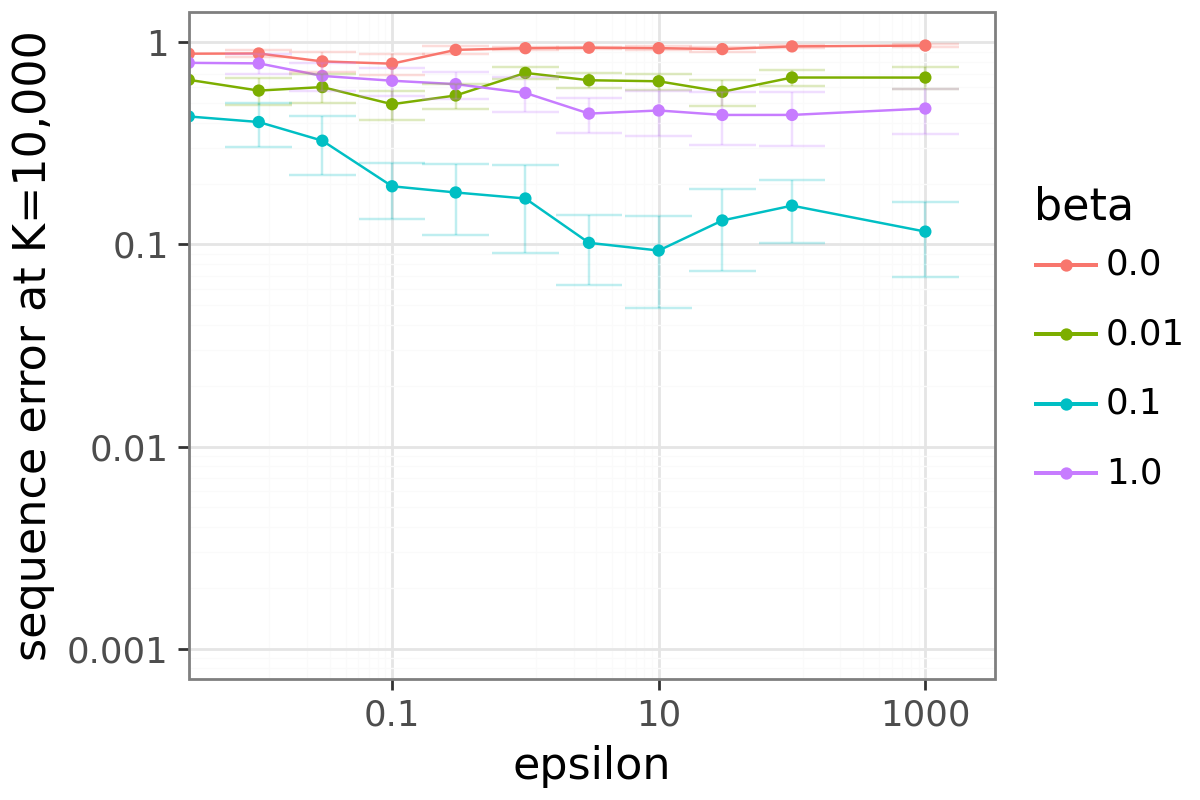}
    \vspace{-2mm}
    \caption{PPO: $\varepsilon$ sweep.}
\end{subfigure}
\hfill
\begin{subfigure}[t]{0.32\columnwidth}
    \includegraphics[width=\linewidth]{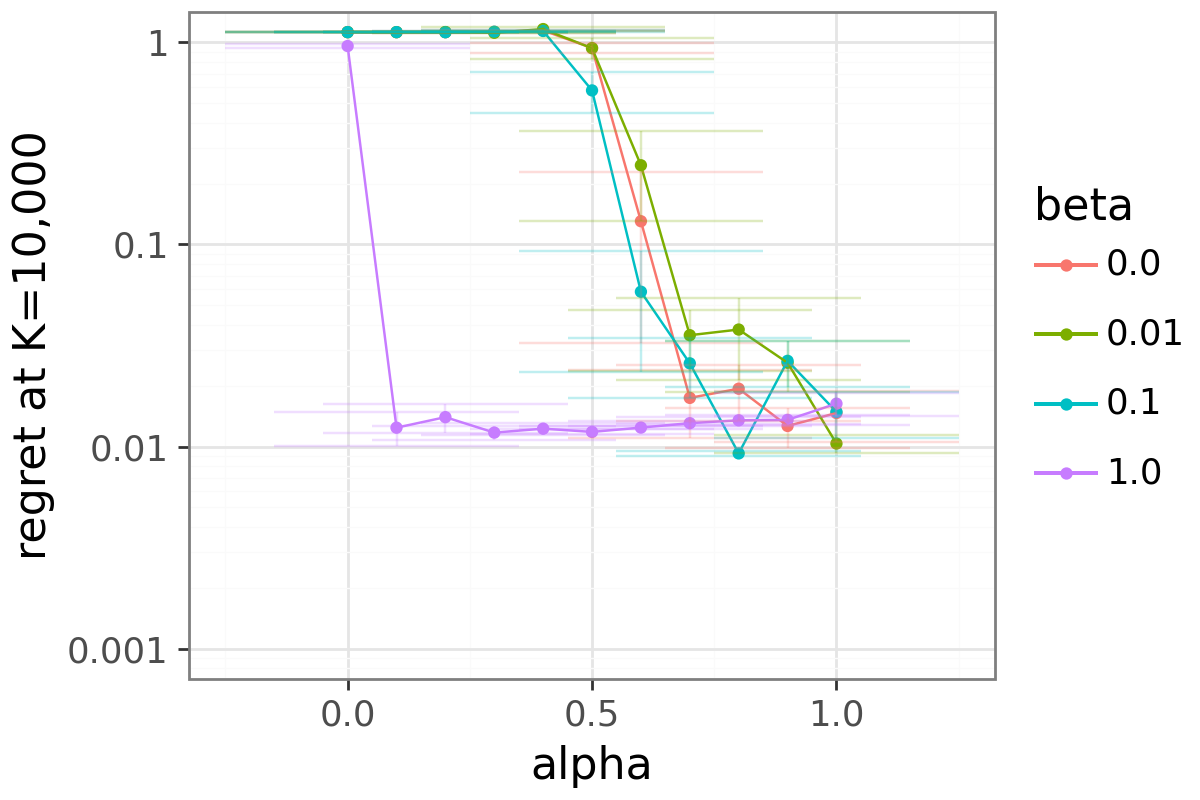}
    \vspace{-2mm}
    \caption{PMPO: $\alpha$ sweep.}
\end{subfigure}
\vspace{-2mm}
\caption{Hyperparameter sensitivity under combined friction ($H{=}5$, all frictions active), 10 seeds.}
\label{fig:combo_tune}
\vspace{-2mm}
\end{figure}

\paragraph{Scaling to complex domains (Section~\ref{sec:combo}).}
Figure~\ref{fig:combo_tune} shows the sweep at $H{=}5$ with staleness, actor bugs, reward corruption, and rare discovery active at their single-friction operating points.

\subsection{Reward Shaping Sensitivity}

\begin{figure}[ht!]
\centering
\vspace{-2mm}
\begin{subfigure}[t]{0.48\columnwidth}
    \includegraphics[width=\linewidth]{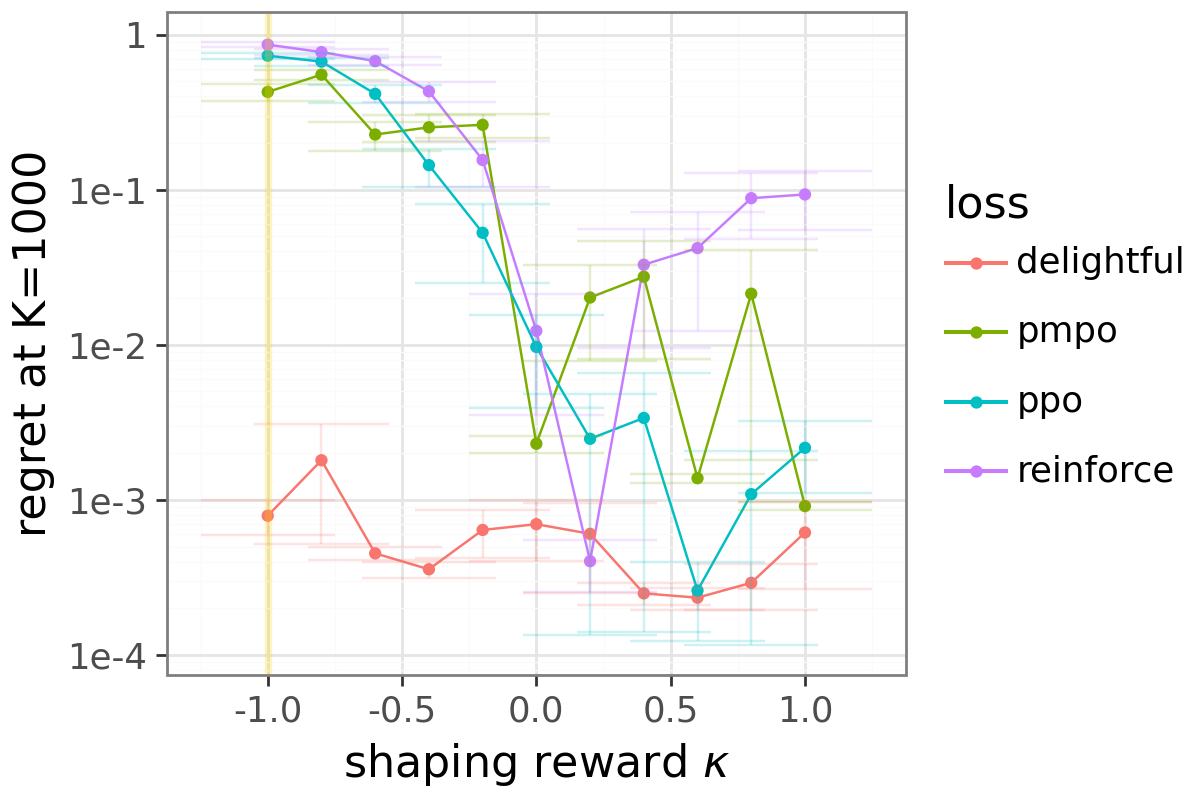}
    \vspace{-3mm}
    \caption{Sequence error at $K{=}1000$ vs.\ $\kappa$.}
    \label{fig:explore_kappa_final}
\end{subfigure}
\hfill
\begin{subfigure}[t]{0.48\columnwidth}
    \includegraphics[width=\linewidth]{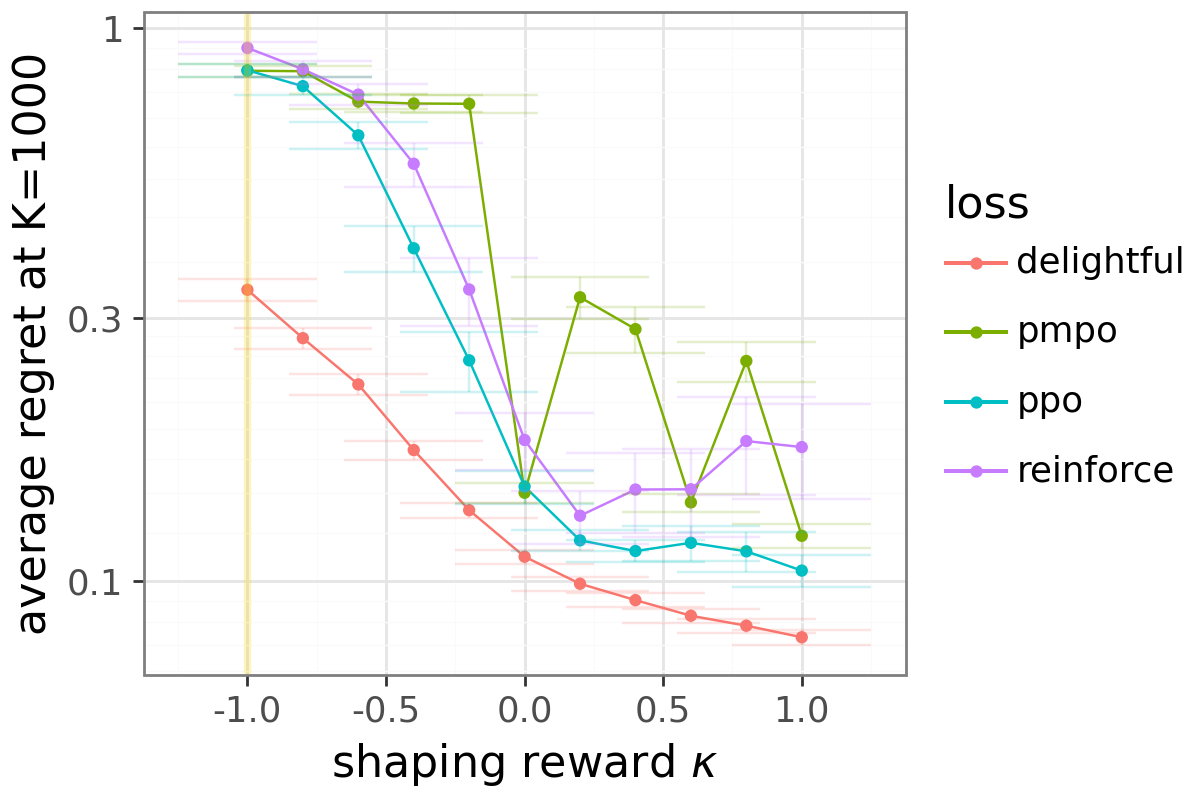}
    \vspace{-3mm}
    \caption{Average sequence error vs.\ $\kappa$.}
    \label{fig:explore_kappa_ave}
\end{subfigure}
\vspace{-2mm}
\caption{Sensitivity to reward shaping under rare discovery ($p_C{=}10^{-3}$, $H{=}5$).
DG outperforms the baselines for every $\kappa$, with the largest gains in the hedonic trap; plot axes label sequence error as regret.}
\label{fig:explore_kappa}
\vspace{-2mm}
\end{figure}

To test whether DG's advantage depends on the reward structure, we sweep the shaping parameter $\kappa \in [-1, 1]$ with oracle discovery rate $p_C{=}10^{-3}$ and $H{=}5$.
Figure~\ref{fig:explore_kappa} shows that DG outperforms baselines across all values of $\kappa$, with the largest gains in the hedonic trap regime ($\kappa < 0$) where partial progress is penalized.

The gap persists across the hyperparameter settings tested.

\subsection{Practical Considerations}
\label{app:practical}

For deployment beyond the controlled diagnostics studied here, we recommend the following.
Normalize advantages within each prompt group before forming delight, so that the gate scale does not depend on reward magnitude.
Cap surprisal at a fixed threshold (e.g.\ $\ell_{\max}=20$) to avoid extreme gates from out-of-distribution actions.
Monitor high-delight samples for reward-hacking or verifier errors; the Kondo analysis of \citet{osband2026doesgradientsparkjoy} identifies false delight as the primary failure mode.
When reliable behavior probabilities are available, DG can be combined with clipped ratios or V-trace-style correction as complementary safeguards.
An adaptive temperature $\eta$ proportional to the running standard deviation of $U\ell$ is a natural extension but was not needed in our experiments.

\newpage
\newpage

\section*{NeurIPS Paper Checklist}

\begin{enumerate}

\item {\bf Claims}
    \item[] Question: Do the main claims made in the abstract and introduction accurately reflect the paper's contributions and scope?
    \item[] Answer: \answerYes{}
    \item[] Justification: The abstract and introduction state four contributions: (1) identifying negative learning from high-surprisal failures as a failure mode (validated theoretically in Section~4 and empirically in Section~3), (2) proving that sign-aware surprisal gating suppresses contaminated perpendicular failure noise while preserving rare successes (Propositions~1--2), (3) showing that exact behavior probabilities do not yield the best finite-batch update under staleness (Section~3), and (4) introducing controlled distributed-friction diagnostics (Section~5). The abstract states that all experiments are controlled diagnostics, and the conclusion (Section~7) states the limitations.
    \item[] Guidelines:
    \begin{itemize}
        \item The answer \answerNA{} means that the abstract and introduction do not include the claims made in the paper.
        \item The abstract and/or introduction should clearly state the claims made, including the contributions made in the paper and important assumptions and limitations. A \answerNo{} or \answerNA{} answer to this question will not be perceived well by the reviewers. 
        \item The claims made should match theoretical and experimental results, and reflect how much the results can be expected to generalize to other settings. 
        \item It is fine to include aspirational goals as motivation as long as it is clear that these goals are not attained by the paper. 
    \end{itemize}

\item {\bf Limitations}
    \item[] Question: Does the paper discuss the limitations of the work performed by the authors?
    \item[] Answer: \answerYes{}
    \item[] Justification: Section~7 (Conclusion) states three limitations: the experiments are controlled diagnostics with no end-to-end asynchronous system or inference-stack comparison; DG is deliberately biased and the theory characterizes that bias only in the analyzed bandit; and \emph{false delight} (high-surprisal positive-advantage samples from reward variance, verifier error, or reward hacking) is the main failure mode, with its operating boundary from prior work. It also recommends practical mitigations.
    \item[] Guidelines:
    \begin{itemize}
        \item The answer \answerNA{} means that the paper has no limitation while the answer \answerNo{} means that the paper has limitations, but those are not discussed in the paper. 
        \item The authors are encouraged to create a separate ``Limitations'' section in their paper.
        \item The paper should point out any strong assumptions and how robust the results are to violations of these assumptions (e.g., independence assumptions, noiseless settings, model well-specification, asymptotic approximations only holding locally). The authors should reflect on how these assumptions might be violated in practice and what the implications would be.
        \item The authors should reflect on the scope of the claims made, e.g., if the approach was only tested on a few datasets or with a few runs. In general, empirical results often depend on implicit assumptions, which should be articulated.
        \item The authors should reflect on the factors that influence the performance of the approach. For example, a facial recognition algorithm may perform poorly when image resolution is low or images are taken in low lighting. Or a speech-to-text system might not be used reliably to provide closed captions for online lectures because it fails to handle technical jargon.
        \item The authors should discuss the computational efficiency of the proposed algorithms and how they scale with dataset size.
        \item If applicable, the authors should discuss possible limitations of their approach to address problems of privacy and fairness.
        \item While the authors might fear that complete honesty about limitations might be used by reviewers as grounds for rejection, a worse outcome might be that reviewers discover limitations that aren't acknowledged in the paper. The authors should use their best judgment and recognize that individual actions in favor of transparency play an important role in developing norms that preserve the integrity of the community. Reviewers will be specifically instructed to not penalize honesty concerning limitations.
    \end{itemize}

\item {\bf Theory assumptions and proofs}
    \item[] Question: For each theoretical result, does the paper provide the full set of assumptions and a complete (and correct) proof?
    \item[] Answer: \answerYes{}
    \item[] Justification: All formal results (Lemma~1, Propositions~1--2) state their assumptions explicitly (softmax policy, $K$-armed bandit, contaminated sampling model $\mu=(1-\rho)\pi+\rho\nu$, reward structure). Complete proofs appear in Appendix~B. Remarks~1--2 in the appendix provide supplementary bounds and the exact importance-weighting connection.
    \item[] Guidelines:
    \begin{itemize}
        \item The answer \answerNA{} means that the paper does not include theoretical results. 
        \item All the theorems, formulas, and proofs in the paper should be numbered and cross-referenced.
        \item All assumptions should be clearly stated or referenced in the statement of any theorems.
        \item The proofs can either appear in the main paper or the supplemental material, but if they appear in the supplemental material, the authors are encouraged to provide a short proof sketch to provide intuition. 
        \item Inversely, any informal proof provided in the core of the paper should be complemented by formal proofs provided in appendix or supplemental material.
        \item Theorems and Lemmas that the proof relies upon should be properly referenced. 
    \end{itemize}

    \item {\bf Experimental result reproducibility}
    \item[] Question: Does the paper fully disclose all the information needed to reproduce the main experimental results of the paper to the extent that it affects the main claims and/or conclusions of the paper (regardless of whether the code and data are provided or not)?
    \item[] Answer: \answerYes{}
    \item[] Justification: Appendix~A provides full MNIST details (architecture, optimizer, learning rate, baseline, staleness model). Appendix~C provides full token reversal details (architecture, optimizer, batch structure, grouped baseline, complete hyperparameter grids). All methods share architecture, optimizer, and baseline; they differ only in gradient weighting.
    \item[] Guidelines:
    \begin{itemize}
        \item The answer \answerNA{} means that the paper does not include experiments.
        \item If the paper includes experiments, a \answerNo{} answer to this question will not be perceived well by the reviewers: Making the paper reproducible is important, regardless of whether the code and data are provided or not.
        \item If the contribution is a dataset and\slash or model, the authors should describe the steps taken to make their results reproducible or verifiable. 
        \item Depending on the contribution, reproducibility can be accomplished in various ways. For example, if the contribution is a novel architecture, describing the architecture fully might suffice, or if the contribution is a specific model and empirical evaluation, it may be necessary to either make it possible for others to replicate the model with the same dataset, or provide access to the model. In general. releasing code and data is often one good way to accomplish this, but reproducibility can also be provided via detailed instructions for how to replicate the results, access to a hosted model (e.g., in the case of a large language model), releasing of a model checkpoint, or other means that are appropriate to the research performed.
        \item While NeurIPS does not require releasing code, the conference does require all submissions to provide some reasonable avenue for reproducibility, which may depend on the nature of the contribution. For example
        \begin{enumerate}
            \item If the contribution is primarily a new algorithm, the paper should make it clear how to reproduce that algorithm.
            \item If the contribution is primarily a new model architecture, the paper should describe the architecture clearly and fully.
            \item If the contribution is a new model (e.g., a large language model), then there should either be a way to access this model for reproducing the results or a way to reproduce the model (e.g., with an open-source dataset or instructions for how to construct the dataset).
            \item We recognize that reproducibility may be tricky in some cases, in which case authors are welcome to describe the particular way they provide for reproducibility. In the case of closed-source models, it may be that access to the model is limited in some way (e.g., to registered users), but it should be possible for other researchers to have some path to reproducing or verifying the results.
        \end{enumerate}
    \end{itemize}

\item {\bf Open access to data and code}
    \item[] Question: Does the paper provide open access to the data and code, with sufficient instructions to faithfully reproduce the main experimental results, as described in supplemental material?
    \item[] Answer: \answerYes{}
    \item[] Justification: Code is available at \url{https://github.com/google-deepmind/egg}. All experimental details are also provided in Appendices~A and~C.
    \item[] Guidelines:
    \begin{itemize}
        \item The answer \answerNA{} means that paper does not include experiments requiring code.
        \item Please see the NeurIPS code and data submission guidelines (\url{https://neurips.cc/public/guides/CodeSubmissionPolicy}) for more details.
        \item While we encourage the release of code and data, we understand that this might not be possible, so \answerNo{} is an acceptable answer. Papers cannot be rejected simply for not including code, unless this is central to the contribution (e.g., for a new open-source benchmark).
        \item The instructions should contain the exact command and environment needed to run to reproduce the results. See the NeurIPS code and data submission guidelines (\url{https://neurips.cc/public/guides/CodeSubmissionPolicy}) for more details.
        \item The authors should provide instructions on data access and preparation, including how to access the raw data, preprocessed data, intermediate data, and generated data, etc.
        \item The authors should provide scripts to reproduce all experimental results for the new proposed method and baselines. If only a subset of experiments are reproducible, they should state which ones are omitted from the script and why.
        \item At submission time, to preserve anonymity, the authors should release anonymized versions (if applicable).
        \item Providing as much information as possible in supplemental material (appended to the paper) is recommended, but including URLs to data and code is permitted.
    \end{itemize}

\item {\bf Experimental setting/details}
    \item[] Question: Does the paper specify all the training and test details (e.g., data splits, hyperparameters, how they were chosen, type of optimizer) necessary to understand the results?
    \item[] Answer: \answerYes{}
    \item[] Justification: Section~3 and Appendix~A specify MNIST details (architecture, optimizer, learning rate sweep, baseline). Section~5 and Appendix~C specify token reversal details including architecture, optimizer (Adam, lr=$10^{-4}$), batch structure, grouped baseline, hyperparameter grids, tuning protocol (10 tuning seeds, 30 evaluation seeds), and compute requirements.
    \item[] Guidelines:
    \begin{itemize}
        \item The answer \answerNA{} means that the paper does not include experiments.
        \item The experimental setting should be presented in the core of the paper to a level of detail that is necessary to appreciate the results and make sense of them.
        \item The full details can be provided either with the code, in appendix, or as supplemental material.
    \end{itemize}

\item {\bf Experiment statistical significance}
    \item[] Question: Does the paper report error bars suitably and correctly defined or other appropriate information about the statistical significance of the experiments?
    \item[] Answer: \answerYes{}
    \item[] Justification: All experiments report $\pm 1$ standard error of the mean across independent seeds: 30 seeds for MNIST and token reversal, 100 seeds for bandits. Error bars are visible in all figures and explicitly identified in figure captions (e.g., Figures~1--8). The factor of variability is the random seed (controlling initialization and data sampling).
    \item[] Guidelines:
    \begin{itemize}
        \item The answer \answerNA{} means that the paper does not include experiments.
        \item The authors should answer \answerYes{} if the results are accompanied by error bars, confidence intervals, or statistical significance tests, at least for the experiments that support the main claims of the paper.
        \item The factors of variability that the error bars are capturing should be clearly stated (for example, train/test split, initialization, random drawing of some parameter, or overall run with given experimental conditions).
        \item The method for calculating the error bars should be explained (closed form formula, call to a library function, bootstrap, etc.)
        \item The assumptions made should be given (e.g., Normally distributed errors).
        \item It should be clear whether the error bar is the standard deviation or the standard error of the mean.
        \item It is OK to report 1-sigma error bars, but one should state it. The authors should preferably report a 2-sigma error bar than state that they have a 96\% CI, if the hypothesis of Normality of errors is not verified.
        \item For asymmetric distributions, the authors should be careful not to show in tables or figures symmetric error bars that would yield results that are out of range (e.g., negative error rates).
        \item If error bars are reported in tables or plots, the authors should explain in the text how they were calculated and reference the corresponding figures or tables in the text.
    \end{itemize}

\item {\bf Experiments compute resources}
    \item[] Question: For each experiment, does the paper provide sufficient information on the computer resources (type of compute workers, memory, time of execution) needed to reproduce the experiments?
    \item[] Answer: \answerYes{}
    \item[] Justification: Appendix~C states that all experiments run on a single CPU machine, each token reversal experiment completes in under two minutes, and the full experimental campaign requires approximately 200 CPU-hours total.
    \item[] Guidelines:
    \begin{itemize}
        \item The answer \answerNA{} means that the paper does not include experiments.
        \item The paper should indicate the type of compute workers CPU or GPU, internal cluster, or cloud provider, including relevant memory and storage.
        \item The paper should provide the amount of compute required for each of the individual experimental runs as well as estimate the total compute. 
        \item The paper should disclose whether the full research project required more compute than the experiments reported in the paper (e.g., preliminary or failed experiments that didn't make it into the paper). 
    \end{itemize}
    
\item {\bf Code of ethics}
    \item[] Question: Does the research conducted in the paper conform, in every respect, with the NeurIPS Code of Ethics \url{https://neurips.cc/public/EthicsGuidelines}?
    \item[] Answer: \answerYes{}
    \item[] Justification: The research involves algorithmic improvements to policy gradient methods, tested on synthetic tasks (MNIST, bandits, token reversal). No human subjects, private data, or dual-use concerns are involved.
    \item[] Guidelines:
    \begin{itemize}
        \item The answer \answerNA{} means that the authors have not reviewed the NeurIPS Code of Ethics.
        \item If the authors answer \answerNo, they should explain the special circumstances that require a deviation from the Code of Ethics.
        \item The authors should make sure to preserve anonymity (e.g., if there is a special consideration due to laws or regulations in their jurisdiction).
    \end{itemize}

\item {\bf Broader impacts}
    \item[] Question: Does the paper discuss both potential positive societal impacts and negative societal impacts of the work performed?
    \item[] Answer: \answerNA{}
    \item[] Justification: This is foundational research on policy gradient methods, tested on synthetic tasks. It is not tied to a particular application. The method improves the robustness of distributed RL training, which is broadly applicable; we do not foresee specific negative societal impacts beyond those inherent to any improvement in RL training efficiency.
    \item[] Guidelines:
    \begin{itemize}
        \item The answer \answerNA{} means that there is no societal impact of the work performed.
        \item If the authors answer \answerNA{} or \answerNo, they should explain why their work has no societal impact or why the paper does not address societal impact.
        \item Examples of negative societal impacts include potential malicious or unintended uses (e.g., disinformation, generating fake profiles, surveillance), fairness considerations (e.g., deployment of technologies that could make decisions that unfairly impact specific groups), privacy considerations, and security considerations.
        \item The conference expects that many papers will be foundational research and not tied to particular applications, let alone deployments. However, if there is a direct path to any negative applications, the authors should point it out. For example, it is legitimate to point out that an improvement in the quality of generative models could be used to generate Deepfakes for disinformation. On the other hand, it is not needed to point out that a generic algorithm for optimizing neural networks could enable people to train models that generate Deepfakes faster.
        \item The authors should consider possible harms that could arise when the technology is being used as intended and functioning correctly, harms that could arise when the technology is being used as intended but gives incorrect results, and harms following from (intentional or unintentional) misuse of the technology.
        \item If there are negative societal impacts, the authors could also discuss possible mitigation strategies (e.g., gated release of models, providing defenses in addition to attacks, mechanisms for monitoring misuse, mechanisms to monitor how a system learns from feedback over time, improving the efficiency and accessibility of ML).
    \end{itemize}
    
\item {\bf Safeguards}
    \item[] Question: Does the paper describe safeguards that have been put in place for responsible release of data or models that have a high risk for misuse (e.g., pre-trained language models, image generators, or scraped datasets)?
    \item[] Answer: \answerNA{}
    \item[] Justification: The paper does not release models, datasets, or pretrained weights. The contribution is an algorithmic modification to policy gradients.
    \item[] Guidelines:
    \begin{itemize}
        \item The answer \answerNA{} means that the paper poses no such risks.
        \item Released models that have a high risk for misuse or dual-use should be released with necessary safeguards to allow for controlled use of the model, for example by requiring that users adhere to usage guidelines or restrictions to access the model or implementing safety filters. 
        \item Datasets that have been scraped from the Internet could pose safety risks. The authors should describe how they avoided releasing unsafe images.
        \item We recognize that providing effective safeguards is challenging, and many papers do not require this, but we encourage authors to take this into account and make a best faith effort.
    \end{itemize}

\item {\bf Licenses for existing assets}
    \item[] Question: Are the creators or original owners of assets (e.g., code, data, models), used in the paper, properly credited and are the license and terms of use explicitly mentioned and properly respected?
    \item[] Answer: \answerYes{}
    \item[] Justification: MNIST is a standard public dataset cited via \citet{lecun1998gradient}. All referenced algorithms (PPO, PMPO, IMPALA, etc.) are properly cited.
    \item[] Guidelines:
    \begin{itemize}
        \item The answer \answerNA{} means that the paper does not use existing assets.
        \item The authors should cite the original paper that produced the code package or dataset.
        \item The authors should state which version of the asset is used and, if possible, include a URL.
        \item The name of the license (e.g., CC-BY 4.0) should be included for each asset.
        \item For scraped data from a particular source (e.g., website), the copyright and terms of service of that source should be provided.
        \item If assets are released, the license, copyright information, and terms of use in the package should be provided. For popular datasets, \url{paperswithcode.com/datasets} has curated licenses for some datasets. Their licensing guide can help determine the license of a dataset.
        \item For existing datasets that are re-packaged, both the original license and the license of the derived asset (if it has changed) should be provided.
        \item If this information is not available online, the authors are encouraged to reach out to the asset's creators.
    \end{itemize}

\item {\bf New assets}
    \item[] Question: Are new assets introduced in the paper well documented and is the documentation provided alongside the assets?
    \item[] Answer: \answerYes{}
    \item[] Justification: The experimental code is released at \url{https://github.com/google-deepmind/egg} with a README describing how to reproduce each experiment; no new datasets or models are released.
    \item[] Guidelines:
    \begin{itemize}
        \item The answer \answerNA{} means that the paper does not release new assets.
        \item Researchers should communicate the details of the dataset\slash code\slash model as part of their submissions via structured templates. This includes details about training, license, limitations, etc. 
        \item The paper should discuss whether and how consent was obtained from people whose asset is used.
        \item At submission time, remember to anonymize your assets (if applicable). You can either create an anonymized URL or include an anonymized zip file.
    \end{itemize}

\item {\bf Crowdsourcing and research with human subjects}
    \item[] Question: For crowdsourcing experiments and research with human subjects, does the paper include the full text of instructions given to participants and screenshots, if applicable, as well as details about compensation (if any)? 
    \item[] Answer: \answerNA{}
    \item[] Justification: The paper does not involve crowdsourcing or human subjects.
    \item[] Guidelines:
    \begin{itemize}
        \item The answer \answerNA{} means that the paper does not involve crowdsourcing nor research with human subjects.
        \item Including this information in the supplemental material is fine, but if the main contribution of the paper involves human subjects, then as much detail as possible should be included in the main paper. 
        \item According to the NeurIPS Code of Ethics, workers involved in data collection, curation, or other labor should be paid at least the minimum wage in the country of the data collector. 
    \end{itemize}

\item {\bf Institutional review board (IRB) approvals or equivalent for research with human subjects}
    \item[] Question: Does the paper describe potential risks incurred by study participants, whether such risks were disclosed to the subjects, and whether Institutional Review Board (IRB) approvals (or an equivalent approval/review based on the requirements of your country or institution) were obtained?
    \item[] Answer: \answerNA{}
    \item[] Justification: The paper does not involve human subjects research.
    \item[] Guidelines:
    \begin{itemize}
        \item The answer \answerNA{} means that the paper does not involve crowdsourcing nor research with human subjects.
        \item Depending on the country in which research is conducted, IRB approval (or equivalent) may be required for any human subjects research. If you obtained IRB approval, you should clearly state this in the paper. 
        \item We recognize that the procedures for this may vary significantly between institutions and locations, and we expect authors to adhere to the NeurIPS Code of Ethics and the guidelines for their institution. 
        \item For initial submissions, do not include any information that would break anonymity (if applicable), such as the institution conducting the review.
    \end{itemize}

\item {\bf Declaration of LLM usage}
    \item[] Question: Does the paper describe the usage of LLMs if it is an important, original, or non-standard component of the core methods in this research? Note that if the LLM is used only for writing, editing, or formatting purposes and does \emph{not} impact the core methodology, scientific rigor, or originality of the research, declaration is not required.
    \item[] Answer: \answerNA{}
    \item[] Justification: LLMs are not used as a component of the core method. The method is a modification to policy gradient weighting applied to a small causal transformer trained from scratch.
    \item[] Guidelines:
    \begin{itemize}
        \item The answer \answerNA{} means that the core method development in this research does not involve LLMs as any important, original, or non-standard components.
        \item Please refer to our LLM policy in the NeurIPS handbook for what should or should not be described.
    \end{itemize}

\end{enumerate}

\end{document}